\documentclass[manuscript,screen]{acmart}
\DeclareUnicodeCharacter{202F}{\,}
\AtBeginDocument{%
  }

\usepackage{amsthm}

\usepackage{subfigure}
\usepackage{enumitem}
\usepackage{caption}
\usepackage{multirow} 
\usepackage{multicol} 
\usepackage{booktabs}
\usepackage{xcolor}
\usepackage{enumitem}
\usepackage{hyperref}
\usepackage{makecell}

\DeclareMathOperator*{\argmin}{arg\,min}
\DeclareMathOperator*{\argmax}{arg\,max}

\setcopyright{acmlicensed}
\copyrightyear{2018}
\acmYear{2018}
\acmDOI{XXXXXXX.XXXXXXX}
\acmConference[XXX]{X}{X}{X}
\acmISBN{978-1-4503-XXXX-X/2018/06}

\begin{document}

\title{Generative Model Unlearning: A Survey through Target Events, Unlearning Operators, and Evaluation Protocols}

\author{Xiaohua Feng}
\affiliation{%
  \institution{Zhejiang University}
  \city{Hangzhou}
  \country{China}
}

\author{Jiaming Zhang}
\affiliation{%
  \institution{Zhejiang University}
  \city{Hangzhou}
  \country{China}
}

\author{Fengyuan Yu}
\affiliation{%
  \institution{Zhejiang University}
  \city{Hangzhou}
  \country{China}
}

\author{Chengye Wang}
\affiliation{%
  \institution{Zhejiang University}
  \city{Hangzhou}
  \country{China}
}

\author{Li Zhang}
\affiliation{%
  \institution{Zhejiang University}
  \city{Hangzhou}
  \country{China}
}

\author{Kaixiang Li}
\affiliation{%
  \institution{Hangzhou Dianzi University}
  \city{Hangzhou}
  \country{China}}

\author{Yuyuan Li}
\affiliation{%
  \institution{Hangzhou Dianzi University}
  \city{Hangzhou}
  \country{China}}

\author{Lingjuan Lyu}
\affiliation{%
  \institution{Sony Research}
  \city{Hangzhou}
  \country{Japan}}

\author{Chaochao Chen}
\authornote{Corresponding Author.}
\affiliation{%
  \institution{Zhejiang University}
  \city{Hangzhou}
  \country{China}
}

\author{Jianwei Yin}
\authornotemark[1]
\affiliation{%
  \institution{Zhejiang University}
  \city{Hangzhou}
  \country{China}
}

\renewcommand{\shortauthors}{Xiaohua Feng et al.}

\begin{abstract}
With the rapid advancement of generative models, privacy, copyright, safety, and reliability risks have attracted growing attention.
To mitigate these risks, machine unlearning has been increasingly adapted from traditional classification models to generative settings.
Despite notable progress, existing studies remain fragmented in their target definitions, unlearning mechanisms, and evaluation protocols, making objective comparison difficult across models, modalities, and applications.
Moreover, modality-specific surveys often overlook the shared structure of Generative Model Unlearning (GenMU).
To address this gap, we provide a comprehensive review of GenMU and formulate it as target-constrained distributional projection: given a target event, an unlearning operator transforms the generative distribution to suppress target-related outputs while preserving useful behavior and controlling operator cost.
Under this framework, an unlearning request is specified by a target event, implemented through an unlearning operator, and assessed by empirical evidence over target suppression, distribution preservation, and operator cost.
We further reorganize existing GenMU studies under this view. 
From this perspective, we provide the first explicit and unified account of how GenMU connects to mainstream applications, including privacy protection, copyright and style protection, safety alignment, hallucination mitigation, and deployment defense.
Finally, we identify key open problems and future directions toward reliable, scalable, robust, and auditable GenMU.
We consistently maintain the related open-source materials at \href{https://github.com/caxLee/Generative-model-unlearning-survey}{https://github.com/caxLee/GenMU-Survey}.
\end{abstract}



\received{20 February 2007}
\received[revised]{12 March 2009}
\received[accepted]{5 June 2009}

\maketitle

\section{Introduction}
\label{sec:intro}

\begin{figure*}[t]
  \centering
  \includegraphics[width=\linewidth]{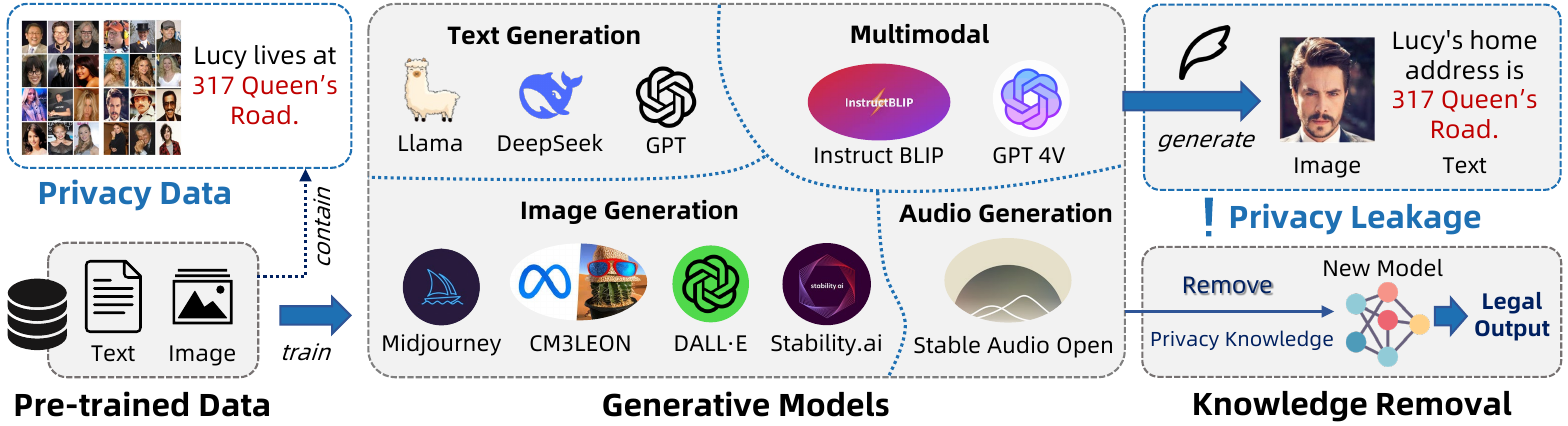}
  \caption{Privacy leakage in GMs. GMs trained on large-scale datasets may inadvertently reproduce specific training data fragments during generation. When such fragments contain sensitive information, they pose significant privacy risks. To mitigate this, knowledge removal aims to suppress the influence of designated private data from the model after training.}
  \label{fig:intro}
  \vspace{-15pt}
\end{figure*}

Generative Models (GMs) have become a foundational technology for modern artificial intelligence, supporting text generation~\cite{huang2023language}, image synthesis~\cite{ho2020denoising}, audio creation~\cite{Forsgren_Martiros_2022}, and multimodal reasoning~\cite{kim2021vilt} across diverse applications.
However, their ability to model high-dimensional data distributions also makes them prone to reproducing sensitive, copyrighted, or unsafe content observed during pretraining or fine-tuning.
Recent studies~\cite{carlini2021extracting,carlini2023extracting} show that such reproduction may expose private information either explicitly, by regenerating original data fragments, or implicitly, by revealing sensitive associations through paraphrases, semantic cues, or cross-modal prompts.
As illustrated in Fig.~\ref{fig:intro}, this risk turns knowledge removal from an optional safety mechanism into a necessary requirement for privacy protection~\cite{kuner2020eu,kuppa2021towards}, copyright compliance, and responsible deployment under regulations such as the General Data Protection Regulation (GDPR)~\cite{voigt2017eu}.

These concerns naturally motivate Generative Model Unlearning (GenMU), whose goal is to suppress designated target knowledge from a trained GM while preserving non-target generative behavior.
A straightforward solution is to retrain the model after deleting the target data, but this is computationally prohibitive for modern GMs and often infeasible in real deployment.
More importantly, retraining equivalence, which serves as a reference notion in classical machine unlearning, becomes insufficient in generative settings.
GMs produce stochastic, open-ended, and semantically structured outputs, so successful unlearning cannot be judged only by parameter changes or by performance on a fixed test set.
Instead, it must ask whether the post-unlearning model avoids target-related generations under diverse prompts, semantic transformations, and modality-specific queries while maintaining useful behavior elsewhere.

Recent progress has produced a wide range of GenMU methods~\cite{jang2023knowledge,li2024machine,feng2025controllable,li2024single,li2026conesep,li2026habit,li2026retrack,fu2026air,li2026tema}, including gradient-based unlearning, retain-set regularization, preference optimization, representation editing, retrieval modification, and inference-time guidance.
Although these methods demonstrate the feasibility of post-hoc knowledge suppression, their rapid development has also exposed a more fundamental problem: existing studies differ in what they define as the target, how they transform the GM, and which evidence they use to claim successful unlearning.
Consequently, empirical results are difficult to compare because different works may operate on different target structures, rely on different access assumptions, and report metrics that estimate different aspects of unlearning.

This fragmentation can be traced to three missing links in the current understanding of GenMU.
\textit{First, the \textbf{target of unlearning} is rarely specified as a structured event.}
A target may correspond to a memorized instance~\cite{jang2023knowledge,kong2023data,feng2025controllable}, an identity~\cite{choi2024opt,mason-williams2025machine,kim2025do}, an artistic style~\cite{eldan2023s,zhang2024forget,gandikota2023erasing}, an entity relation~\cite{yang2024cliperase}, or an input-conditioned behavior~\cite{yu2023unlearning,yao2024large}.
Without an explicit description of the target event, including its target query scope, preservation scope, and target presence criterion, it is difficult to determine where the target may appear, what should be preserved, how target presence should be judged, and where over-unlearning begins.
\textit{Second, unlearning methods are often compared without clarifying the \textbf{unlearning operator} they implement.}
Some methods reshape the training objective~\cite{jang2023knowledge,feng2024fine,buarbulescu2024each,wang2025selective}, some steer preferences~\cite{zhang2024negative,fan2024simplicity,choi2024opt,mekala2025alternate}, some edit localized representations~\cite{wu2023depn,jia2025wagle,hu2025falcon,dou2025avoiding}, some manipulate model-difference directions~\cite{ilharco2023editing,gao2024ethos,liu2024towards}, and others control prompts, retrieval contexts, or decoding distributions~\cite{pawelczyk2024context,li2024get,huang2024offset}.
These methods may all reduce target-related outputs, but they rely on different assumptions and induce different reachable distribution families.
\textit{Third, \textbf{evaluation results} are often reported without a shared interpretation.}
Metrics such as extraction likelihood~\cite{jang2023knowledge,carlini2021extracting}, classifier scores~\cite{gandikota2023erasing}, similarity measures~\cite{maini2024tofu}, toxicity rates~\cite{ni2024forgetting,liu2024towards}, generation quality~\cite{feng2025controllable}, and attack success rates~\cite{zhang2024generate} provide useful evidence, but they are often used to support different claims about target suppression, distribution preservation, or operator cost.
Without mapping them into a unified evaluation framework, empirical comparisons can be misleading.

Existing reviews of GenMU are organized by generative domain, covering text~\cite{zhang2024right,liu2025rethinking,barez2025open}, image~\cite{kim2025comprehensive}, audio, and multimodal generation~\cite{liu2025mllmubench}, with Liu et al.~\cite{liu2024machine} providing a broader cross-domain account.
While these efforts offer useful coverage, a modality-centered organization has a structural limitation: it describes where unlearning methods are applied, but does not reveal what makes unlearning problems share the same structure across modalities.
We argue that this shared structure is captured by two modality-independent dimensions.
The first is the target event: text paraphrases, image style shifts, and cross-modal prompts look different on the surface, but they all instantiate the same underlying question of how far the target query scope must extend for suppression to hold, making target-event specification a domain-agnostic design choice.
The second is the unlearning operator: objective tuning, representation editing, and decoding guidance appear to be modality-specific implementations, but they are in fact different mechanisms for transforming the model's conditional generative distribution, a transformation that is regardless of modality.
A survey on GenMU should therefore go beyond aligning modality-specific literatures, and instead identify these shared structural elements, namely the target event, the unlearning operator, and the evaluation protocols, that jointly determine what a GenMU problem is, how it is solved, and how success is verified.

To this end, we formulate GenMU as a \textit{target-constrained distributional projection} problem.
Given a trained conditional distribution and a target unlearning event, GenMU seeks a transformed distribution that suppresses target-related generations while controlling distribution preservation and operator cost.
From this view, we organize GenMU into three coupled components.
i) The target event specifies the target query scope, the preservation scope, and the target presence criterion.
ii) The unlearning operator defines the transformation and determines the corresponding reachable distribution family.
iii) The evaluation protocols provide evidence for three goals: whether the target is effectively suppressed under standard and transformed queries, whether broad retain and target-adjacent behavior are preserved, and whether the unlearning operator is practically feasible.
Under this formulation, modality is not the primary organizing axis.
Instead, text, image, audio, and multimodal generation instantiate the same framework through different query scopes, preservation scopes, presence criteria, operators, and empirical estimators.
We further establish a unified connection between GenMU and mainstream applications, showing that privacy protection, copyright and style protection, safety alignment, hallucination mitigation, and deployment defense can be viewed as concrete instantiations of target events and unlearning operators.
Finally, we identify the remaining open problems and future directions under this framework, including formal guarantees, target boundary specification, robust black-box and continual unlearning, cross-modal certification, and benchmark standardization.

\begin{figure*}[t]
  \centering
  \includegraphics[width=\linewidth]{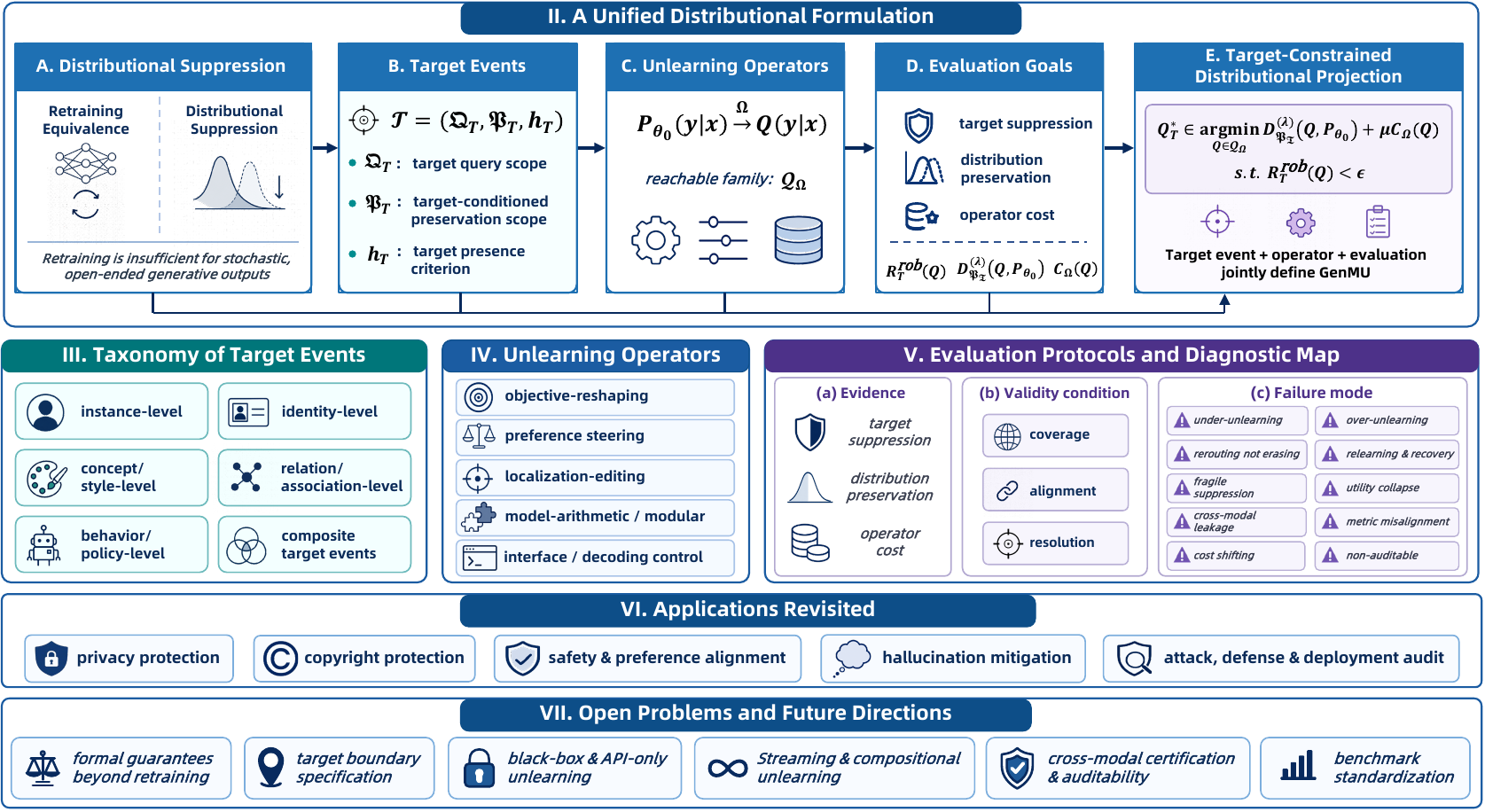}
  \caption{The proposed analytical framework of GenMU. We formulate GenMU as a target-constrained distributional projection and analyze existing studies through target events, unlearning operators, and evaluation protocols.}
  \label{fig:framework}
  \vspace{-15pt}
\end{figure*}

Based on this view, we make the following contributions: 
\begin{itemize}[leftmargin=*] \setlength{\itemsep}{2pt}
    \item \textbf{A unified analytical framework for GenMU.}
    We formulate GenMU as target-constrained distributional projection and organize the field through three coupled components: target events, unlearning operators, and evaluation protocols.
    Under this unified analytical framework, existing studies can be fairly compared in terms of the content to be suppressed, the way the model distribution is transformed, and the evidence used to support the unlearning claim.

    \item \textbf{A unified connection between GenMU and mainstream applications.}
    To the best of our knowledge, this survey is the first to explicitly connect GenMU with privacy protection, copyright and style protection, safety alignment, hallucination mitigation, and deployment defense under a common analytical framework.
    We show that these applications are not separate taxonomies, but concrete instantiations of target events and unlearning operators within GenMU.

    \item \textbf{A framework-driven analysis of open problems and future directions.}
    We identify the remaining gaps exposed by our proposed framework, including formal guarantees beyond retraining equivalence, target boundary specification, robust black-box and continual unlearning, cross-modal certification and auditability, and benchmark standardization.
    This analysis clarifies which parts of the GenMU pipeline remain underdeveloped and outlines future directions toward reliable, scalable, robust, and auditable unlearning.
\end{itemize}
As summarized in Fig.~\ref{fig:framework}, the rest of this survey follows the same logical chain.
Section~\ref{sec:foun} first introduces the unified distributional formulation.
Sections~\ref{sec:target_taxonomy} and~\ref{sec:operator_taxonomy} then specify the two structural components of GenMU, namely what should be suppressed and how the model distribution is transformed.
Section~\ref{sec:evaluation} further explains how unlearning claims are supported by empirical evaluation evidence and how the limitations of existing evaluations can be diagnosed through failure modes.
Sections~\ref{sec:applications} and~\ref{sec:future} finally revisit mainstream applications as framework instantiations and discuss the open problems and future directions revealed by this analysis.
By shifting the focus from modality-wise enumeration to target events, unlearning operators, and evaluation evidence, this survey aims to provide a coherent foundation for reliable, scalable, and auditable GenMU.
\section{A Unified Distributional Formulation}
\label{sec:foun}
The introduction frames GenMU as a target-constrained transformation of a trained generative distribution.
This section formalizes this view with a consistent set of concepts.
We first explain why classical retraining equivalence is insufficient for GenMU.
We then define two structural components: the target event and the unlearning operator.
Building on them, we specify the distributional goals of evaluation.
Finally, we formulate GenMU as a target-constrained distributional projection problem, which serves as the basis for the later target-event taxonomy, operator taxonomy, and evaluation protocol.

\subsection{From Retraining Equivalence to Distributional Suppression}
\label{sec:from_retraining_to_suppression}

Classical machine unlearning~\cite{bourtoule2021machine,heng2023machine} is commonly formulated through retraining equivalence.
Let $\mathcal{D}_0$ be the original training dataset, $\mathcal{D}_f \subset \mathcal{D}_0$ the data to be removed, and $\mathcal{D}_r=\mathcal{D}_0\setminus \mathcal{D}_f$ the retained dataset.
Given a training algorithm $\mathcal{A}$, the original model is $\boldsymbol{\theta}_0=\mathcal{A}(\mathcal{D}_0)$, while an ideal unlearned model is expected to behave like a model retrained on $\mathcal{D}_r$.
Thus, a classical unlearning algorithm $\mathcal{U}$ produces $\boldsymbol{\theta}^*=\mathcal{U}(\boldsymbol{\theta}_0,\mathcal{D}_0,\mathcal{D}_f)$, and aims to satisfy
\begin{equation}
    \mathrm{Dis}\bigl(
    \mathrm{Pr}(\mathcal{U}(\boldsymbol{\theta}_0,\mathcal{D}_0,\mathcal{D}_f))
    \|
    \mathrm{Pr}(\mathcal{A}(\mathcal{D}_r))
    \bigr)
    \leq \epsilon ,
    \label{eq:classical_unlearning}
\end{equation}
where $\mathrm{Dis}(\cdot\|\cdot)$ denotes a distance between distributions over model parameters.

Although Eq.~\eqref{eq:classical_unlearning} is conceptually clear, it does not directly capture the central difficulty of GenMU.
First, retraining modern GMs is usually computationally prohibitive, making $\mathcal{A}(\mathcal{D}_r)$ unavailable as a practical reference.
Second, GMs define stochastic and open-ended conditional distributions, so parameter-level similarity to a retrained model is neither necessary nor sufficient for suppressing target-related generations.
Third, target information may persist through semantic neighbors, paraphrases, associations, or cross-modal cues even after exact memorized fragments disappear~\cite{ko2024boosting}.
Therefore, GenMU should not be formulated as retraining equivalence, but as distributional suppression of a target event under explicit constraints on non-target behavior and practical cost.


\subsection{Target Events}
\label{sec:target_events}

The first structural component of GenMU is the target event.
A target event specifies where the target may appear, how target presence is identified, and how target-related content is separated from retainable content.
Let $T$ denote the target content to be suppressed, such as a sample, identity, concept, relation, or behavior, and let $\mathcal{T}$ denote its event-level specification in the generative space.
Formally, we define
\begin{equation}
    \mathcal{T} = (\mathfrak{Q}_T, \mathfrak{P}_T, h_T),
    \label{eq:target_event}
\end{equation}
where $\mathfrak{Q}_T$ is the target query scope, $\mathfrak{P}_T$ is the target-conditioned preservation scope, and $h_T:\mathcal{X}\times\mathcal{Y}_m\rightarrow[0,1]$ is the target presence criterion.
The query scope $\mathfrak{Q}_T$ is induced by a base query distribution $\Pi_T$ and a transformation set $\mathcal{A}_T$, 
\begin{equation}
    \mathfrak{Q}_T
    =
    \{a_{\#}\Pi_T: a\in\mathcal{A}_T\},
    \label{eq:target_query_scope}
\end{equation}
where $a_{\#}\Pi_T$ denotes the transformed query distribution.
Here, $\Pi_T$ captures standard target-triggering conditions, while $\mathcal{A}_T$ captures transformations under which suppression should remain valid.
The preservation scope $\mathfrak{P}_T$ plays the analogous role on the non-target side.
Let $\Pi_R$ denote a broad retain query distribution, and let $\mathcal{W}_T$ be a set of non-negative target-conditioned focus weights.
For any $w\in\mathcal{W}_T$, define the reweighted retain distribution $d\Pi_R^w(x) = \frac{w(x)d\Pi_R(x)}{\mathbb{E}_{x\sim\Pi_R}[w(x)]}$.
%
%
The preservation scope is then
\begin{equation}
    \mathfrak{P}_T
    =
    \{\Pi_R^w:w\in\mathcal{W}_T\}.
    \label{eq:preservation_scope}
\end{equation}
This construction includes broad retain preservation and boundary preservation as two canonical slices.
When $w_0(x)\equiv1$, $\Pi_R^{w_0}=\Pi_R$ recovers the broad retain distribution.
When $w_{\partial T}$ concentrates mass on target-adjacent but retainable queries, $\Pi_R^{w_{\partial T}}$ recovers the target boundary distribution, denoted by $\Pi_{\partial T} := \Pi_R^{w_{\partial T}}$.
Finally, the criterion $h_T$ defines target presence for each input-output pair.

\subsection{Unlearning Operators}
\label{sec:conditional_distribution_operator}

We model a trained GM as a conditional distribution $P_{\boldsymbol{\theta}_0}(y|x)$, where $x\in\mathcal{X}$ denotes the input condition and $y\in\mathcal{Y}_m$ the generated output.
A GenMU method transforms $P_{\boldsymbol{\theta}_0}$ into a post-unlearning distribution $Q(y|x)$ through an unlearning operator $\Omega$.
The operator $\Omega$ induces a reachable distribution family $\mathcal{Q}_{\Omega}$, consisting of all attainable post-unlearning distributions under its assumptions, access conditions, and computational constraints.
Thus, GenMU can be studied as a constrained transformation:
\begin{equation}
    P_{\boldsymbol{\theta}_0}(y|x) \xrightarrow{\Omega}Q(y|x).
    \label{eq:operator_transformation}
\end{equation}
This view separates the mechanism of unlearning from its implementation.
A parameter update~\cite{jang2023knowledge,li2024machine,li2024single}, a representation edit~\cite{eldan2023s,zhang2024forget,cheng2024multidelete}, a retrieval modification, a prompt transformation, and a decoding-time adjustment may all induce a valid post-unlearning distribution $Q$.
What matters for analysis is the reachable distribution family $\mathcal{Q}_{\Omega}$ determined by the unlearning operator.
This family specifies which transformations are feasible and therefore determines how a method can trade off target suppression, distribution preservation, and operator cost.
Section~\ref{sec:operator_taxonomy} develops the operator taxonomy.

\subsection{Evaluation Goals}
\label{sec:evaluation_groups}

The third component of the framework is evaluation.
Given a target event $\mathcal{T}$, an unlearning operator $\Omega$, and a post-unlearning distribution $Q\in\mathcal{Q}_{\Omega}$, evaluation asks whether the transformation from $P_{\boldsymbol{\theta}_0}$ to $Q$ satisfies three goals: target suppression, distribution preservation, and operator cost.

\paragraph{Target suppression}
Target suppression asks whether target-related generations are reduced under the target query scope.
At the distributional level, this goal is represented by the target risk
\begin{equation}
    R_T(Q)
    =
    \mathbb{E}_{x\sim \Pi_T}
    \mathbb{E}_{y\sim Q(\cdot|x)}
    \bigl[
    h_T(x,y)
    \bigr],
    \label{eq:target_risk}
\end{equation}
and its robust counterpart
\begin{equation}
    R_T^{\mathrm{rob}}(Q)
    =
    \sup_{\Pi\in\mathfrak{Q}_T}
    \mathbb{E}_{x\sim \Pi}
    \mathbb{E}_{y\sim Q(\cdot|x)}
    \bigl[
    h_T(x,y)
    \bigr].
    \label{eq:robust_target_risk}
\end{equation}
The former evaluates suppression under standard target queries, while the latter evaluates suppression over the target query scope.

\paragraph{Distribution preservation}
Distribution preservation asks whether the transformation preserves non-target behavior.
At the distributional level, it is evaluated over the preservation scope $\mathfrak{P}_T$.
For any slice $\Pi_R^w\in\mathfrak{P}_T$, define
\begin{equation}
    D_{T,w}(Q,P_{\boldsymbol{\theta}_0})
    =
    \mathbb{E}_{x\sim \Pi_R^w}
    \bigl[
    d(Q(\cdot|x),P_{\boldsymbol{\theta}_0}(\cdot|x))
    \bigr],
    \label{eq:preservation_slice_distortion}
\end{equation}
where $d(\cdot,\cdot)$ is a distance between conditional output distributions.
Different preservation effects are captured by different slices of $\mathfrak{P}_T$, rather than by different distance functions.
%
%
Then, the broad retain slice is obtained by $w_0(x)\equiv1$: $D_{\mathrm{ret}}(Q,P_{\boldsymbol{\theta}_0})=D_{T,w_0}(Q,P_{\boldsymbol{\theta}_0})$.
%
%
The target-adjacent boundary slice is obtained by $w_{\partial T}$: $B_T(Q,P_{\boldsymbol{\theta}_0})=D_{T,w_{\partial T}}(Q,P_{\boldsymbol{\theta}_0})$.
%
%
Thus, retain distortion and boundary distortion are not two unrelated quantities.
They are two projections of the same preservation scope $\mathfrak{P}_T$: $D_{\mathrm{ret}}$ measures broad retain preservation, while $B_T$ measures preservation on the target-adjacent slice where over-unlearning is most likely to occur.

\paragraph{Operator cost}
Operator cost asks whether the transformation is feasible under practical constraints.
We denote the cost of realizing $Q$ under $\Omega$ by $C_{\Omega}(Q)$. 
This term may include training time, required data, memory, parameter storage, inference overhead, and access conditions.
These goals define what successful GenMU means at the distributional level.

\subsection{GenMU as Target-Constrained Distributional Projection}
\label{sec:distributional_projection}

We now combine the target event, the unlearning operator, and the evaluation goals into a formulation.
Given the original distribution $P_{\boldsymbol{\theta}_0}$, a target event $\mathcal{T}$, and a reachable distribution family $\mathcal{Q}_{\Omega}$, define the preservation functional over $\mathfrak{P}_T$ as
\begin{equation}
    D_{\mathfrak{P}_T}^{(\lambda)}
    (Q,P_{\boldsymbol{\theta}_0})
    =
    D_{\mathrm{ret}}(Q,P_{\boldsymbol{\theta}_0})
    +
    \lambda B_T(Q,P_{\boldsymbol{\theta}_0}),
    \label{eq:preservation_functional}
\end{equation}
where $\lambda\geq0$ controls the relative emphasis on the target-adjacent boundary slice.
GenMU seeks
%
\begin{equation}
    Q_T^* \in \operatorname*{arg\,min}_{Q\in\mathcal{Q}_{\Omega}}
    D_{\mathfrak{P}_T}^{(\lambda)}(Q,P_{\boldsymbol{\theta}_0})
    +
    \mu C_{\Omega}(Q), \quad \mathrm{s.t.} \quad
    R_T^{\mathrm{rob}}(Q)\leq \epsilon,
    \label{eq:genmu_projection}
\end{equation}
where $\mu\geq0$ controls the relative importance of operator cost.
The constraint represents target suppression over $\mathfrak{Q}_T$.
The first objective term represents distribution preservation over the unified preservation scope $\mathfrak{P}_T$.
Its broad retain and target-adjacent boundary effects are recovered by the slices $w_0$ and $w_{\partial T}$.
The second objective term represents operator cost.
Thus, Eq.~\eqref{eq:genmu_projection} gives a unified interpretation of GenMU.
\section{Taxonomy of Target Events}
\label{sec:target_taxonomy}

This section builds a taxonomy of target events $\mathcal{T} = (\mathfrak{Q}_T, \mathfrak{P}_T, h_T)$ by examining how their three functional parts are instantiated in GenMU.
Rather than classifying targets by modality or application domain, we focus on the event structures that determine what should be suppressed, how suppression is tested, and where over-unlearning begins.
Although target events may appear in text, image, audio, or multimodal outputs, their category is defined by the structure of $\mathcal{T}$.
We identify five types: instance-level, identity-level, concept/style-level, relation/association-level, and behavior/policy-level events.
They differ in whether the target is a concrete sample, an identity, an open semantic attribute, an entity association, or an input-conditioned behavior; for each type, the target query scope, target presence criterion, and target-conditioned preservation scope are instantiated differently.
A comparison is provided in Tab.~\ref{tab:target_event_atlas}.

\subsection{Instance-Level Target Events}
\label{sec:instance_target_events}

An instance-level target event corresponds to one or a few protected samples, such as a private sentence~\cite{jang2023knowledge}, training image~\cite{kong2023data,li2024machine,feng2025controllable}, copyrighted audio clip, confidential fragment, or image-text pair~\cite{yang2024cliperase}.
Its key property is that the target is anchored to concrete instances rather than an open semantic class.
Let $\mathcal{Y}_T=\{y_i\}_{i=1}^{n}$ denote the protected instances.
The target query scope $\mathfrak{Q}_T$ contains queries that may elicit them, typically induced by a base query distribution $\Pi_T$ and transformations $\mathcal{A}_T$ such as paraphrases, partial contexts, prompt variants, viewpoint shifts, and cross-modal reconstruction.
The target presence criterion can be defined as
\begin{equation}
    h_T(x,y)
    =
    \max_{y_i\in\mathcal{Y}_T}
    k_m(y,y_i),
    \label{eq:instance_presence}
\end{equation}
where $k_m$ is a modality-dependent similarity function.
The preservation scope $\mathfrak{P}_T$ contains broad retain samples and a target-adjacent boundary slice $\Pi_{\partial T}=\Pi_R^{w_{\partial T}}$ that concentrates on similar but retainable content.
Thus, the main challenge is to suppress protected instances without damaging nearby non-target samples~\cite{li2024single}.

\subsection{Identity-Level Target Events}
\label{sec:identity_target_events}

An identity-level target event covers all generated expressions of a specific identity, such as a person, speaker~\cite{mason-williams2025machine,kim2025do}, face identity, user profile~\cite{choi2024opt}, or private subject.
Unlike instance-level events, it is not tied to one sample, since the same identity may appear through names, descriptions, faces, voices, contexts, or modalities.
Let $e$ denote the target identity.
The target query scope $\mathfrak{Q}_T$ contains queries that may refer to or elicit $e$, expanded by transformations across aliases, descriptions, poses, voices, and modalities.
The target presence criterion is
\begin{equation}
    h_T(x,y)
    =
    s_{\mathrm{id}}(y;e),
    \label{eq:identity_presence}
\end{equation}
where $s_{\mathrm{id}}$ may be an identity recognizer, speaker verifier~\cite{kim2025do}, entity linker, multimodal detector, or human/LLM judge.
The preservation scope $\mathfrak{P}_T$ contains broad non-target identities and a boundary slice $\Pi_{\partial T}$ that concentrates on similar but non-target identities or attributes.
Thus, identity-level unlearning must suppress the target identity without suppressing nearby identities or retainable attributes.

\subsection{Concept- and Style-Level Target Events}
\label{sec:concept_style_target_events}

A concept- or style-level target event corresponds to an open semantic attribute, visual category, artistic style, toxic concept, unsafe content type, or other non-instance pattern~\cite{eldan2023s,zhang2024forget,gandikota2023erasing,kumari2023ablating,huang2024receler}.
Its key property is that the target forms an open set rather than a finite sample collection or a single identity.
The target query scope contains queries likely to elicit the concept or style, including transformed expressions such as paraphrases, euphemisms, prompt substitutions, style transfer, composition, and cross-modal reasoning.
The target presence criterion is usually
\begin{equation}
    h_T(x,y)
    =
    s_{\mathrm{con}}(x,y;c),
    \label{eq:concept_presence}
\end{equation}
where $c$ is the target concept or style and $s_{\mathrm{con}}$ may be a classifier, toxicity scorer~\cite{adolphs2023cringe,yao2024large,lu2022quark}, object detector, CLIP-like similarity model~\cite{yang2024cliperase}, style recognizer, safety model~\cite{schramowski2023safe}, or human/LLM evaluator.
Since concepts and styles are often ambiguous, $h_T$ is better viewed as a graded signal.
The preservation scope $\mathfrak{P}_T$ contains broad retainable concepts and a boundary slice $\Pi_{\partial T}$ that concentrates on neighboring concepts, styles, or attributes that should remain available.
Thus, this target requires robust suppression while controlling over-suppression near the boundary~\cite{lu2024mace,lee2025localized}.

\subsection{Relation- and Association-Level Target Events}
\label{sec:relation_association_target_events}

A relation- or association-level target event corresponds to a specific relation between entities, attributes, or modalities.
The goal is to suppress the association, not the endpoints themselves.
Let $a$ and $b$ denote two entities or attributes, and let $r(a,b)$ denote the target relation.
The target query scope contains queries that may elicit this relation, including paraphrased, reverse, indirect, multi-hop, or cross-modal queries.
The target presence criterion is
\begin{equation}
    h_T(x,y)
    =
    s_{\mathrm{rel}}(x,y;r(a,b)).
    \label{eq:relation_presence}
\end{equation}
where $s_{\mathrm{rel}}$ may be implemented by relation extraction, factual QA, entity linking, image-text matching~\cite{yang2024cliperase}, caption consistency, or human/LLM judgment.
The preservation scope $\mathfrak{P}_T$ contains retainable knowledge about the entities and a boundary slice $\Pi_{\partial T}$ that concentrates on endpoints and neighboring relations.
Thus, relation-level unlearning must break the target association without deleting the endpoints.

\newcommand{\cmark}{$\checkmark$}
\newcommand{\smark}{$\star$}
\newcommand{\dash}{--}

\begin{table*}[t]
\centering
\caption{Structure of target events in GenMU. $\checkmark$ denotes the defining structural anchor of a basic target-event type;
$\star$ denotes composition over one or more basic target events.}
\label{tab:target_event_atlas}
\scriptsize
\setlength{\tabcolsep}{3.2pt}
\renewcommand{\arraystretch}{1.12}
\resizebox{\linewidth}{!}{
\begin{tabular}{p{0.17\linewidth} ccccc p{0.14\linewidth} p{0.20\linewidth} p{0.18\linewidth} p{0.18\linewidth}}
\toprule
\multirow{2}{*}{\textbf{Target Event}}
& \multicolumn{5}{c}{\textbf{Event Structure}}
& \multirow{2}{*}{\textbf{Anchor}}
& \multirow{2}{*}{\textbf{Query Scope $\boldsymbol{\mathfrak{Q}_T}$}}
& \multirow{2}{*}{\textbf{Presence Criterion $\boldsymbol{h_T}$}}
& \multirow{2}{*}{\textbf{Preservation Scope $\boldsymbol{\mathfrak{P}_T}$}} \\
\cmidrule(lr){2-6}
& \textbf{Sample} & \textbf{Entity} & \textbf{Open} & \textbf{Rel.} & \textbf{Beh.}
& & & & \\
\midrule

Instance-level
& \cmark & \dash & \dash & \dash & \dash
& $\{y_i\}$
& Direct / partial / reconstruction
& $\max_i k_m(y,y_i)$
& Near retainable samples \\

Identity-level
& \dash & \cmark & \dash & \dash & \dash
& $e$
& Names / aliases / faces / voices
& $s_{\mathrm{id}}(y;e)$
& Similar non-target identities \\

Concept/style-level
& \dash & \dash & \cmark & \dash & \dash
& $c$
& Prompts / paraphrases / style shifts
& $s_{\mathrm{con}}(x,y;c)$
& Neighboring concepts or styles \\

Relation/association-level
& \dash & \dash & \dash & \cmark & \dash
& $r(a,b)$
& Forward / reverse / multi-hop
& $s_{\mathrm{rel}}(x,y;r(a,b))$
& Endpoints and nearby relations \\

Behavior/policy-level
& \dash & \dash & \dash & \dash & \cmark
& $\pi$
& Triggering / jailbreak / adversarial
& $s_{\mathrm{beh}}(x,y;\pi)$
& Benign adjacent requests \\

\midrule
Composite target events
& \smark & \smark & \smark & \smark & \smark
& $\mathcal{T}=\{\mathcal{T}_k\}_{k=1}^{K}$
& Union / weighted query scopes
& $\{h_{T_k}\}_{k=1}^{K}$
& Event-wise preservation scopes \\

\bottomrule
\end{tabular}
}

\vspace{-8pt}
\end{table*}

\subsection{Behavior- and Policy-Level Target Events}
\label{sec:behavior_policy_target_events}

A behavior- or policy-level target event corresponds to an input-conditioned response pattern rather than a fixed output object.
Examples include harmful instruction following~\cite{chakraborty2024crossmodal}, hallucination, biased responses~\cite{yu2023unlearning,yao2024large}, unsafe tool use~\cite{cheng2025tool}, privacy-violating assistance, refusal failures, or preference misalignment~\cite{zhang2024negative,mekala2025alternate,fan2024simplicity}.
Its key property is that target presence depends jointly on the input and output.
The target query scope contains prompts or contexts likely to trigger the behavior, including jailbreaks, role-play prompts, obfuscation, translation, multi-turn contexts, indirect instructions, and adversarial settings~\cite{pawelczyk2024context,muresanu2024unlearnable,thaker2024guardrail,bhaila2025soft}.
The target presence criterion is
\begin{equation}
    h_T(x,y)
    =
    s_{\mathrm{beh}}(x,y;\pi),
    \label{eq:behavior_presence}
\end{equation}
where $\pi$ denotes the target behavior or policy violation.
The function $s_{\mathrm{beh}}$ may be a safety classifier~\cite{schramowski2023safe,liu2024large,ren2025general}, hallucination detector, factual verifier, tool-use checker~\cite{cheng2025tool}, reward model~\cite{wu2024universal}, preference model~\cite{zhang2024negative,choi2024opt,park2024direct}, or human/LLM evaluator.
The preservation scope $\mathfrak{P}_T$ contains broad benign behavior and a boundary slice $\Pi_{\partial T}$ that concentrates on benign but target-adjacent requests.
Thus, behavior- and policy-level events require suppressing undesirable behaviors without making the model overly conservative~\cite{ji2024reversing,sanyal2025alu}.

\subsection{Composite Target Events}
\label{sec:composite_target_events}

Many practical GenMU requests are composite.
For example, copyright-related requests~\cite{tian2024forget,li2024single}, privacy-related requests~\cite{mason-williams2025machine,kim2025do}, safety-related requests~\cite{schramowski2023safe,cheng2025tool}, or alignment-related requests~\cite{lu2022quark,zhang2024negative,choi2024opt} may involve multiple target event types.
Thus, an application scenario should not be forced into a single category.
We decompose a composite request into $\mathcal{T}=\{\mathcal{T}_1,\mathcal{T}_2,\ldots,\mathcal{T}_K\}$.
Each target event $\mathcal{T}_k=(\mathfrak{Q}_{T_k},h_{T_k},\mathfrak{P}_{T_k})$ specifies its own query scope, presence criterion, and preservation scope.
The overall robust target risk can be evaluated by the worst event,
%
\begin{equation}
    R_{\mathcal{T}}^{\mathrm{rob}}(Q)
    =
    \max_{1\leq k\leq K}
    R_{T_k}^{\mathrm{rob}}(Q),
    \;
    \text{or by a weighted aggregation,}
    \;
    R_{\mathcal{T}}^{\mathrm{rob}}(Q)
    =
    \sum_{k=1}^{K}
    \alpha_k R_{T_k}^{\mathrm{rob}}(Q),
    \;
    \alpha_k\geq0,
    \;
    \sum_{k=1}^{K}\alpha_k=1.
    \label{eq:weighted_composite_robust_risk}
\end{equation}
This decomposition separates application labels from target event categories and makes failures easier to localize.
\section{Unlearning Operators}
\label{sec:operator_taxonomy}

This section addresses how an unlearning method transforms the original distribution $P_{\boldsymbol{\theta}_0}(y|x)$ into a post-unlearning distribution $Q(y|x)$ for a given target event $\mathcal{T}$.
Following Section~\ref{sec:foun}, we formulate an unlearning operator $\Omega$ as the mechanism that induces this transformation and determines the reachable distribution family $\mathcal{Q}_{\Omega}$.
Thus, we organize GenMU methods by their induced distributional transformations rather than by modality, application scenario, or surface-level update scope.
This unlearning operator definition complements the target-event taxonomy: a target event defines where the target may appear, how it is detected, and where its boundary lies, whereas an operator defines how the model distribution is changed in response.
The same target event may be handled by different operators, and the same operator may apply across different targets and modalities.
Their interaction determines the evaluation emphasis among target suppression, distribution preservation, and operator cost.
Accordingly, this section introduces operators as reachable distribution families, analyzes their compatibility with target events, and discusses their evaluation implications.

\subsection{Unlearning Operator as Reachable Distribution Families}
\label{sec:operator_families}

The central distinction among unlearning methods lies in how they restrict the feasible post-unlearning distributions.
We identify five operator families: objective-reshaping operators, preference or contrastive steering operators, localization-editing operators, model-arithmetic or modular operators, and interface or decoding-control operators.
Tab.~\ref{tab:operator_atlas} summarizes each operator family through its reachable distribution family $\mathcal{Q}_{\Omega}$, the key assumption under which the family is meaningful, and the induced distributional mechanism.
Detailed method lists can be found in Appendix~\ref{app:operator_catalog}.

\newcommand{\Qobj}{$\{P_{\boldsymbol{\theta}}:\boldsymbol{\theta}\in\argmin_{\boldsymbol{\theta}}\mathcal{L}_{\mathrm{ER}}(\boldsymbol{\theta};\mathcal{T})\}$}
\newcommand{\Qpref}{$\{Q_{\boldsymbol{\phi}}:\boldsymbol{\phi}\in\argmax_{\boldsymbol{\phi}}\mathcal{J}_{\mathrm{PC}}(\boldsymbol{\phi};\mathcal{T})\}$}
\newcommand{\Qloc}{$\{P_{\boldsymbol{\theta}_0+\Delta}:\Delta\in\mathcal{S}_{\mathcal{T}}\}$}
\newcommand{\Qmod}{$\{P_{\boldsymbol{\theta}_0+M_{\boldsymbol{\phi}}-\sum_k\alpha_k\boldsymbol{v}_k}:M_{\boldsymbol{\phi}}\in\mathcal{M}_{\mathcal{T}},\boldsymbol{v}_k\in\mathcal{V}_{\mathcal{T}}\}$}
\newcommand{\Qint}{$\{Q:Q(y|x)\propto P_{\boldsymbol{\theta}_0}(y|\phi_{\mathcal{T}}(x))\exp[-\eta s_{\mathcal{T}}(x,y)]\}$}

\begin{table*}[t]
\centering
\caption{Unlearning operators in GenMU. $\mathcal{L}_{\mathrm{ER}}$: erase-retain objective; $\mathcal{J}_{\mathrm{PC}}$: preference or contrastive objective. The table defines operator families by their reachable distribution families only; evaluation emphasis is determined by the pair $(\mathcal{T},\Omega)$.}
\label{tab:operator_atlas}
\scriptsize
\setlength{\tabcolsep}{4.2pt}
\renewcommand{\arraystretch}{1.16}
\resizebox{\linewidth}{!}{
\begin{tabular}{llll}
\toprule
\textbf{Unlearning Operator}
& $\boldsymbol{\mathcal{Q}_{\Omega}}$ \textbf{ Reachable Distribution Family}
& \textbf{Key Assumption}
& \textbf{Mechanism} \\
\midrule

Objective-reshaping
& \Qobj
& Objective covers $\mathcal{T}$
& Optimize erase-retain objectives \\

Preference / contrastive
& \Qpref
& Comparisons cover boundary
& Steer relative likelihoods \\

Localization-editing
& \Qloc
& $\mathcal{S}_{\mathcal{T}}$ separates target and retain behavior
& Edit target-related subspace \\

Model-arithmetic / modular
& \Qmod
& Directions or modules are reusable
& Compose vectors or modules \\

Interface / decoding-control
& \Qint
& Control layer covers query scope
& Transform query, context, or decoding \\

\bottomrule
\end{tabular}
}
\vspace{-8pt}
\end{table*}

\paragraph{Objective-reshaping operators}
Objective-reshaping operators suppress target events by optimizing an erase-retain objective.
Their reachable distribution family can be written as
\begin{equation}
    \mathcal{Q}_{\Omega_{\mathrm{obj}}}
    =
    \left\{
    P_{\boldsymbol{\theta}}:
    \boldsymbol{\theta}
    \in
    \argmin_{\boldsymbol{\theta}}
    \left[
    \mathcal{L}_{\mathrm{erase}}(\boldsymbol{\theta};\mathcal{T})
    +
    \lambda_R\mathcal{L}_{\mathrm{retain}}(\boldsymbol{\theta};\mathfrak{P}_T)
    \right]
    \right\}.
    \label{eq:objective_operator}
\end{equation}
Here, the erase term supports target suppression, and the retain term supports distribution preservation.
This family covers gradient ascent and its selective variants for text unlearning~\cite{jang2023knowledge,feng2024fine,buarbulescu2024each,wang2025selective}, retain-regularized or distillation-based updates~\cite{wang2023kga,wang2025towards,kim2023towards,fan2023salun}, second-order or geometry-aware corrections~\cite{gu2024second,wei2025provable,bae2023gradient}, diffusion or image-generation objectives~\cite{kong2023data,gandikota2023erasing,wang2025ace,huang2024receler,kim2024race}, and multimodal or audio variants~\cite{kim2025do,chakraborty2024crossmodal,yang2024cliperase,li2024single,huo2025mmunlearner}.
Its strength is flexibility, since the same objective template can be adapted to instance-level, identity-level, concept/style-level, relation/association-level, and behavior/policy-level events.
Its limitation is that target suppression and distribution preservation must be balanced within the same optimization problem.

\paragraph{Preference or contrastive steering operators}
Preference or contrastive steering operators suppress targets by reshaping relative generation likelihoods between target-related and target-free outputs.
Their reachable distribution family can be written as
%
\begin{equation}
    \mathcal{Q}_{\Omega_{\mathrm{pref}}}
    =
    \left\{
    Q_{\boldsymbol{\phi}}:
    \boldsymbol{\phi}
    \in
    \argmax_{\boldsymbol{\phi}}
    \left[
    \mathbb{E}_{(x,y^+,y^-)}
    \bigl(
    \log Q_{\boldsymbol{\phi}}(y^+|x)
    -
    \log Q_{\boldsymbol{\phi}}(y^-|x)
    \bigr)
    -
    \beta D_{\mathfrak{P}_T}^{(\lambda)}(Q_{\boldsymbol{\phi}},P_{\boldsymbol{\theta}_0})
    \right]
    \right\}.
    \label{eq:preference_operator}
\end{equation}
where $Q_{\boldsymbol{\phi}}$ denotes the distribution induced by steering variables $\boldsymbol{\phi}$, $y^-$ is target-related or undesirable, and $y^+$ is target-free or preferred.
This family includes negative preference optimization and its variants~\cite{zhang2024negative,fan2024simplicity,choi2024opt,mekala2025alternate}, contrastive token- or output-level steering~\cite{adolphs2023cringe,lu2022quark}, and paired image or diffusion preference learning~\cite{park2024direct,wallace2024diffusion}.
It is most useful when the target event is easier to express through comparisons than through a single absolute target presence criterion.
Its effectiveness depends on whether the preference data covers the target query scope and the target boundary distribution.

\paragraph{Localization-editing operators}
Localization-editing operators first identify target-related internal components and then edit them.
The edited component may be a neuron set, parameter subset, representation direction, attention map, latent subspace, sparse feature, or cross-modal alignment pathway.
The reachable family is constrained to localized modifications:
\begin{equation}
    \mathcal{Q}_{\Omega_{\mathrm{loc}}}
    =
    \left\{
    P_{\boldsymbol{\theta}_0+\Delta}:
    \Delta\in\mathcal{S}_{\mathcal{T}}
    \right\},
    \label{eq:localization_family}
\end{equation}
where $\mathcal{S}_{\mathcal{T}}$ is the target-related component set or subspace identified by the localization procedure.
Representative methods include attribution-guided parameter or neuron editing~\cite{wu2023depn,jia2025wagle,hu2025falcon,dou2025avoiding}, representation redirection and subspace intervention~\cite{wang2025large,shen2025lunar,dang2025effects,lizzo2024unlearn,farrell2024applying}, diffusion cross-attention or latent localization~\cite{zhang2024forget,lu2024mace,gandikota2024unified,gong2024reliable,basu2024mechanistic,chavhan2025conceptprune}, and multimodal neuron editing~\cite{liu2025modalityaware}.
This family is attractive when the target event is concentrated in identifiable internal structures.
Its key condition is coverage: the localized component must cover the target event across the target query scope while avoiding target-adjacent retainable content.

\paragraph{Model-arithmetic or modular operators}
Model-arithmetic or modular operators represent unlearning through reusable directions, adapters, low-rank modules, safety vectors, or external modules.
Their reachable distribution family can be written as
\begin{equation}
    \mathcal{Q}_{\Omega_{\mathrm{mod}}}
    =
    \left\{
    P_{\boldsymbol{\theta}_0 + M_{\boldsymbol{\phi}} - \sum_{k=1}^{K}\alpha_k\boldsymbol{v}_k}
    :
    \boldsymbol{v}_k\in\mathcal{V}_{\mathcal{T}},\ 
    M_{\boldsymbol{\phi}}\in\mathcal{M}_{\mathcal{T}}
    \right\}.
    \label{eq:modular_family}
\end{equation}
Here, $\boldsymbol{v}_k$ denotes a model-difference direction, $\mathcal{V}_{\mathcal{T}}$ denotes the target-related direction set, $M_{\boldsymbol{\phi}}$ denotes a modular component, and $\mathcal{M}_{\mathcal{T}}$ denotes the target-related module family.
This family includes task-vector negation and model arithmetic~\cite{ilharco2023editing,gao2024ethos,liu2024towards}, safety vectors and expert/anti-expert directions~\cite{zhou2024making,hu2024separate}, and low-rank or adapter-like modular interventions~\cite{cha2025towards,lyu2024onedimensional,bui2024hiding,lee2025localized}.
Its main advantage is amortization: once a direction or module is learned, it can be stored, reused, combined, or removed.
Its central assumption is separability between target-related and retainable behavior in the chosen direction or module space.

\paragraph{Interface or decoding-control operators}
Interface or decoding-control operators suppress target events without necessarily modifying the base model.
They act through input transformation, retrieval modification, prompt control, external memory filtering, logit offset, or decoding-time reweighting.
Their reachable distribution family can be 
%
\begin{equation}
    \mathcal{Q}_{\Omega_{\mathrm{int}}}
    =
    \left\{
    Q:
    Q(y|x)
    =
    P_{\boldsymbol{\theta}_0}(y|\phi_{\mathcal{T}}(x))
    \;
    \text{or}
    \;
    Q(y|x)
    =
    \frac{
    P_{\boldsymbol{\theta}_0}(y|x)\exp(-\eta s_{\mathcal{T}}(x,y))
    }{
    Z_{\mathcal{T}}(x)
    },
    \;
    \phi_{\mathcal{T}}\in\Phi_{\mathcal{T}},
    \;
    s_{\mathcal{T}}\in\mathcal{H}_{\mathcal{T}},
    \;
    \eta\geq0
    \right\}.
    \label{eq:decoding_operator}
\end{equation}
where $\phi_{\mathcal{T}}$ transforms the input or context, $\Phi_{\mathcal{T}}$ is the admissible transformation class, $s_{\mathcal{T}}$ penalizes target-related outputs, $\mathcal{H}_{\mathcal{T}}$ is the admissible penalty-function class, and $Z_{\mathcal{T}}(x)$ normalizes the distribution.
This family includes in-context and prompt-control methods~\cite{pawelczyk2024context,muresanu2024unlearnable,thaker2024guardrail,bhaila2025soft,liu2024large,ren2025general}, retrieval or external-memory modification~\cite{wang2024machine}, embedding or negative-prompt control~\cite{li2024get,wu2024universal,ban2024understanding}, logit-offset and decoding-time guidance~\cite{huang2024offset,ji2024reversing,sanyal2025alu,choi2024breaking,wu2024cross}, and expert/anti-expert or diffusion guidance~\cite{liu2021dexperts,schramowski2023safe,brack2023sega,sridhar2025prompt,li2025detect,cywinski2025saeuron,wang2025precise,wang2023concept,panda2024fast}.
Its strength is applicability under restricted access.
Its limitation is that the underlying model distribution may remain unchanged, so target suppression must be tested over the full query scope.

\subsection{Compatibility with Target Events}
\label{sec:operator_target_compatibility}

Different unlearning operators are compatible with different target events because target events differ in query scope, presence criterion, and boundary distribution.
Tab.~\ref{tab:target_operator_matrix} gives a compact compatibility matrix.
The table does not prescribe a unique operator for each target event.
Instead, it identifies which operator families are natural starting points under different event structures.

\providecommand{\oppri}{$\checkmark\checkmark$}
\providecommand{\opimp}{$\checkmark$}
\providecommand{\opaux}{$\circ$}
\begin{table}[t]
\centering
\caption{Compatibility between target events and unlearning operators. $\checkmark\checkmark$: primary structure-aligned choice;
$\checkmark$: applicable but event-dependent choice;
$\circ$: secondary or deployment-dependent choice.}
\label{tab:target_operator_matrix}
\scriptsize
\setlength{\tabcolsep}{4pt}
\renewcommand{\arraystretch}{1.15}
\resizebox{0.7\linewidth}{!}{
\begin{tabular}{l c c c c c}
\toprule
\textbf{Target Event}
& \textbf{Objective}
& \textbf{Preference}
& \textbf{Localization}
& \textbf{Modular}
& \textbf{Interface} \\
\midrule
Instance-level
& \oppri & \opaux & \opimp & \opaux & \opimp \\
Identity-level
& \opimp & \opaux & \oppri & \opimp & \opaux \\
Concept/style-level
& \oppri & \opimp & \oppri & \opimp & \opimp \\
Relation/association-level
& \opimp & \opimp & \opimp & \opaux & \oppri \\
Behavior/policy-level
& \opimp & \oppri & \opimp & \opimp & \oppri \\
\bottomrule
\end{tabular}
}
\vspace{-8pt}
\end{table}

This matrix shows why no unlearning operator is universally optimal.
For instance-level events, objective-reshaping directly penalizes protected samples, while interface control may be useful when only query-level access is available.
For identity-level events, localization is natural when identity-specific representations can be identified, but robust evaluation remains necessary because identities can be queried through names, faces, voices, and cross-modal cues.
For concept/style-level events, both objective-reshaping and localization are strong candidates, yet boundary preservation is difficult because neighboring concepts or styles may be semantically close.
For relation/association-level events, interface or retrieval control can remove external access paths, but operators must avoid deleting the endpoints.
For behavior/policy-level events, preference steering and interface control are often effective because behaviors are input-conditioned and can be represented through comparative or deployment-time signals.
Thus, operator selection should be guided by the structure of the target event rather than by modality alone.

\subsection{Evaluation Implications of Target-Operator Pairs}
\label{sec:operator_evaluation_implications}

The evaluation emphasis is not determined by the target event or the unlearning operator alone.
A target event $\mathcal{T}$ specifies what should be probed, what counts as target presence, and where the boundary between target-related and retainable content lies.
An unlearning operator $\Omega$ specifies how $P_{\boldsymbol{\theta}_0}$ is transformed into $Q$ and what reachable distribution family $\mathcal{Q}_{\Omega}$ constrains this transformation.
Therefore, the evidence needed to assess an unlearning result should be interpreted as a property of the pair $(\mathcal{T},\Omega)$.
As shown in Tab.~\ref{tab:target_operator_evaluation}, this subsection summarizes how target events and unlearning operators jointly determine the emphasis among target suppression, distribution preservation, and operator cost.

\providecommand{\evalpri}{$\star$}
\providecommand{\evalimp}{$\bullet$}
\providecommand{\evalaux}{$\circ$}
\providecommand{\evalnaf}{--}

\begin{table*}[t]
\centering
\caption{Pair-conditioned evaluation emphasis for target events and unlearning operators.}
\vspace{-5pt}
\label{tab:target_operator_evaluation}
\scriptsize
\setlength{\tabcolsep}{2.1pt}
\renewcommand{\arraystretch}{1.12}
\resizebox{\linewidth}{!}{
\begin{tabular}{l|ccccc|ccccc|ccccc|ccccc|ccccc}
\toprule
\textbf{Target Event}
& \multicolumn{5}{c|}{\textbf{Objective}}
& \multicolumn{5}{c|}{\textbf{Preference}}
& \multicolumn{5}{c|}{\textbf{Localization}}
& \multicolumn{5}{c|}{\textbf{Modular}}
& \multicolumn{5}{c}{\textbf{Interface}} \\
\cmidrule(lr){2-6}
\cmidrule(lr){7-11}
\cmidrule(lr){12-16}
\cmidrule(lr){17-21}
\cmidrule(lr){22-26}
& TR & RTR & RD & BD & OC
& TR & RTR & RD & BD & OC
& TR & RTR & RD & BD & OC
& TR & RTR & RD & BD & OC
& TR & RTR & RD & BD & OC \\
\midrule

Instance-level
& \evalpri & \evalimp & \evalimp & \evalimp & \evalimp
& \evalpri & \evalimp & \evalimp & \evalpri & \evalaux
& \evalpri & \evalimp & \evalimp & \evalpri & \evalpri
& \evalimp & \evalaux & \evalpri & \evalimp & \evalpri
& \evalimp & \evalpri & \evalimp & \evalaux & \evalpri \\

Identity-level
& \evalimp & \evalpri & \evalimp & \evalpri & \evalimp
& \evalimp & \evalpri & \evalimp & \evalimp & \evalaux
& \evalimp & \evalpri & \evalimp & \evalpri & \evalpri
& \evalimp & \evalimp & \evalpri & \evalpri & \evalpri
& \evalimp & \evalpri & \evalimp & \evalpri & \evalimp \\

Concept/style-level
& \evalpri & \evalpri & \evalimp & \evalpri & \evalimp
& \evalimp & \evalpri & \evalimp & \evalpri & \evalaux
& \evalpri & \evalimp & \evalimp & \evalpri & \evalpri
& \evalimp & \evalimp & \evalimp & \evalpri & \evalpri
& \evalimp & \evalpri & \evalimp & \evalpri & \evalpri \\

Relation/association-level
& \evalpri & \evalimp & \evalpri & \evalimp & \evalimp
& \evalpri & \evalimp & \evalimp & \evalpri & \evalaux
& \evalimp & \evalpri & \evalimp & \evalpri & \evalpri
& \evalimp & \evalaux & \evalpri & \evalpri & \evalpri
& \evalimp & \evalpri & \evalpri & \evalimp & \evalpri \\

Behavior/policy-level
& \evalimp & \evalpri & \evalpri & \evalimp & \evalimp
& \evalimp & \evalpri & \evalpri & \evalimp & \evalaux
& \evalimp & \evalpri & \evalimp & \evalpri & \evalpri
& \evalimp & \evalimp & \evalpri & \evalimp & \evalpri
& \evalimp & \evalpri & \evalimp & \evalimp & \evalpri \\

\bottomrule
\end{tabular}
}
\vspace{2pt}

\footnotesize
TR: Target Risk; RTR: Robust Target Risk; RD: Retain Distortion; BD: Boundary Distortion; OC: Operator Cost.
$\star$: primary emphasis; $\bullet$: important; $\circ$: secondary or setting-dependent; --: usually not central.
The table gives typical emphases for $(\mathcal{T},\Omega)$ pairs rather than fixed evaluation rules.
\vspace{-8pt}
\end{table*}

Target suppression is emphasized when the target event provides a broad or easily transformed query scope and the operator may not fully cover this scope.
The target-side factor is the query scope $\mathfrak{Q}_T$ and the presence criterion $h_T$: instance-level events are often anchored to protected samples, whereas identity-level, concept/style-level, and behavior/policy-level events may reappear through aliases, paraphrases, cross-modal cues, or adversarial contexts.
The operator-side factor is the reachable distribution family $\mathcal{Q}_{\Omega}$: objective-reshaping can directly optimize suppression on represented queries, preference or contrastive steering depends on the coverage of paired examples, localization-editing depends on whether the edited component covers the target across transformed queries, and interface or decoding-control may fail outside the controlled channel.
Therefore, standard target risk is most informative when the target is concretely anchored and directly optimized, while Robust Target Risk becomes central when either the target query scope is broad or the operator only controls part of the access path.

Distribution preservation is emphasized when the target event has retainable content close to the target or when the operator perturbs a broad region of the model distribution.
The target-side factor is the target boundary distribution $\Pi_{\partial T}$: identity-level and concept/style-level events require preserving neighboring identities, concepts, or styles; relation/association-level events require preserving the endpoints while suppressing the relation; behavior/policy-level events require preserving benign adjacent requests.
The operator-side factor is the scope of intervention: objective-reshaping may affect broad retain behavior, preference or contrastive steering may over-suppress if negative examples cover boundary cases, localization-editing may damage retainable features if the localized component is not separable, and modular operators depend on whether reusable directions entangle target and retain behavior.
Therefore, retain distortion is central when broad non-target capability or endpoints must remain available, while Boundary Distortion is central when the target lies close to retainable neighboring content.

Operator cost is emphasized when the target event is large, repeatedly updated, compositionally specified, or requires a broad query scope, and when the chosen operator introduces expensive training, localization, data construction, or inference-time control.
The target-side factor includes the size and structure of the target event, such as whether unlearning concerns one instance, many identities, multiple relations, or a composite target set.
The operator-side factor includes the implementation mode: objective-reshaping may require optimization and retain data, preference or contrastive steering may require curated paired evidence, localization-editing may require attribution or intervention search, modular operators may amortize cost across repeated requests, and interface or decoding-control may shift cost to inference.
Therefore, Operator Cost should be reported as a property of the concrete pair $(\mathcal{T},\Omega)$ and its deployment setting, rather than as a fixed property of an operator family alone.

\section{Evaluation Protocols and Diagnostic Map}
\label{sec:evaluation}

After Sections~\ref{sec:target_taxonomy} and~\ref{sec:operator_taxonomy} define what should be suppressed and how the model distribution is transformed, this section asks how an unlearning claim is empirically supported.
We analyze how finite evidence is constructed, what claim each evidence form supports, and which failures remain hidden when evidence is incomplete.
Thus, evaluation is an evidence-design problem conditioned on the target event $T$ and the unlearning operator $\Omega$.
We keep the target event definition unchanged: $\mathcal{T}=(\mathfrak{Q}_T,\mathfrak{P}_T,h_T)$.
In experiments, components in $T$ are instantiated by finite probes rather than redefined.
Let $\mathcal{I}_T =(\widehat{\mathfrak{Q}}_T,\widehat{\mathfrak{P}}_T,\widehat h_T)$ denote the empirical instantiation, where $\widehat{\mathfrak{Q}}_T$ contains standard and transformed target queries, $\widehat h_T$ estimates target presence, and $\widehat{\mathfrak{P}}_T$ contains broad retain and target-adjacent preservation slices.
Given a post-unlearning distribution $Q\in\mathcal{Q}_{\Omega}$, evaluation produces an evidence vector
\begin{equation}
    \widehat{\mathbf{e}}(Q;\mathcal{I}_T,\Omega)
    =
    \bigl(
    \widehat{R}_T,
    \widehat{R}_T^{\mathrm{rob}},
    \widehat{D}_{\mathrm{ret}},
    \widehat{B}_T,
    C_{\Omega}
    \bigr).
    \label{eq:evaluation_evidence_vector}
\end{equation}
The first two entries support target suppression, the next two support distribution preservation, and the last records operator cost.
The key question is not which metric is reported alone, but whether the evidence constructed through $\mathcal{I}_T$ supports the claimed form of GenMU.

\subsection{Target-Suppression Evidence}
\label{sec:target_suppression_implementation}

Target-suppression evidence depends on where the model is queried and how target presence is judged.
The query side instantiates $\widehat{\mathfrak{Q}}_T$, and the criterion side instantiates $\widehat h_T$.
If either side is under-specified, a low suppression score may reflect a narrow probe rather than genuine target suppression.
In practice, target-suppression evidence commonly takes three common query side instantiations.

\paragraph{Direct probing}
Direct probing evaluates the model under canonical target queries.
For instance-level target events, this includes prefixes, partial contexts, protected samples, reconstruction prompts, and memorization tests~\cite{jang2023knowledge,tirumala2022memorization,carlini2021extracting,wang2025selective,lee2024protecting}.
For identity-level target events, it includes names, aliases, face prompts, speaker prompts, or entity descriptions~\cite{kim2025do,liu2025mllmubench}.
For concept/style-level target events, it includes prompts that explicitly mention the concept, style, object, or unsafe attribute~\cite{eldan2023s,ilharco2023editing,yu2023unlearning,zhou2024making,zhang2024unlearncanvas}.
For relation/association-level target events, it includes direct factual questions, entity-pair prompts, or paired image-text queries~\cite{maini2024tofu,shi2024muse,li2024single}.
For behavior/policy-level target events, it includes standard harmful, hallucination-prone, biased, or tool-use prompts~\cite{zou2023universal,ji2024pku,zhou2024making}.
Direct probing is the minimal evidence for target suppression: if $\widehat R_T$ remains high here, the target has not been suppressed under the most direct access path.
Its limitation is that it covers only a narrow slice of $\mathfrak{Q}_T$.

\paragraph{Transformation probing}
Transformation probing expands the query scope by testing whether suppression survives alternative access paths.
In text generation, this includes paraphrases, synonym substitutions, partial clues, reverse questions, multi-hop prompts~\cite{choi2024cross}, role-play prompts, jailbreaks, and adversarial instructions~\cite{zou2023universal}.
In image generation, it includes prompt substitutions, style shifts, viewpoint changes, object compositions, negative prompts, and prompt-template variants~\cite{schramowski2023safe,zhang2024unlearncanvas,ren2024six,moon2024holistic}.
In audio generation, it includes speaker prompts, prosody changes, transcript variations, and acoustic-condition shifts~\cite{kim2025do}.
In multimodal generation, it includes captioning, VQA, image-to-text queries, text-to-image reconstruction, cross-modal retrieval, and visual grounding prompts~\cite{cheng2024mubench,xu2025pebench,liu2025mllmubench,dontsov2025clear}.
This evidence supports robust target suppression rather than standard target suppression.
It tests whether the target event is suppressed as an event, not merely blocked under one surface form.

\paragraph{Recovery-oriented probing}
Recovery-oriented probing tests whether target information remains recoverable after apparent suppression.
Common implementations include prompt search, jailbreak search, probability-level probing, retrieval augmentation, adapter insertion, light post-unlearning updates, and cross-modal reconstruction~\cite{yao2024machine,lu2024eraser,buarbulescu2024each,shi2024muse}.
This evidence is crucial when $\Omega$ may hide access without changing the underlying distribution, as in interface or decoding-control operators.
It is also important for localization-editing and modular operators, where target information may remain in non-edited components or reusable directions.
Recovery-oriented probing separates stable suppression from rerouting, masking, or deployment-time blocking.

Furthermore, the empirical criterion $\widehat h_T$ must match the target-event structure.
For instance-level events, common criteria include exact match, n-gram overlap, Extraction Likelihood and Memorization Accuracy~\cite{jang2023knowledge,tirumala2022memorization}, Extraction Strength~\cite{carlini2021extracting}, membership inference~\cite{yao2024machine}, embedding similarity, perceptual similarity, and reconstruction success.
For identity-level events, they include entity linking, name matching, face recognition, speaker verification~\cite{kim2025do}, identity classifiers, and cross-modal identity matching~\cite{liu2025mllmubench}.
For concept/style-level events, they include toxicity classifiers~\cite{ilharco2023editing}, safety models, object detectors, style recognizers, CLIP-like similarity~\cite{radford2021learning}, LLM judges~\cite{zhou2024making}, and human annotation.
For relation/association-level events, they include factual QA~\cite{maini2024tofu,shi2024muse}, relation extraction, reverse-query accuracy, entity-pair matching, caption consistency, and image-text matching~\cite{li2024single}.
For behavior/policy-level events, they include harmfulness score~\cite{zhou2024making}, jailbreak success rate, refusal correctness, hallucination detection, factual verification, tool-use violation rate, reward-model score, and safety-classifier score~\cite{ji2024pku}.
The metric name is less important than its alignment with $h_T$: a narrow criterion may miss transformed leakage, while an overly broad one may misclassify target-adjacent retainable outputs as target presence.

\subsection{Distribution-Preservation Evidence}
\label{sec:distribution_preservation_implementation}

Target suppression alone is insufficient because GenMU also requires non-target behavior to remain available.
Distribution-preservation evidence depends on where preservation is measured.
The preservation scope $\widehat{\mathfrak P}_T$ contains two canonical slices: a broad retain slice and a target-adjacent boundary slice.
The former measures average non-target preservation, while the latter tests whether unlearning damages content close to the target.
Conflating these slices often leads to misleading evaluation.

\paragraph{Broad retain preservation}
Broad retain evidence evaluates whether the model remains generally useful after unlearning.
In text generation, this is commonly measured by perplexity, BLEU~\cite{papineni2002bleu}, ROUGE~\cite{lin2004rouge}, METEOR~\cite{banerjee2005meteor}, BERTScore~\cite{zhang2020bertscore}, QA accuracy~\cite{maini2024tofu,shi2024muse,choi2024cross}, factuality scores, diversity metrics such as unique token ratio~\cite{liu2024large} and Rep-4~\cite{sinha2024unstar}, and general benchmarks such as MMLU~\cite{hendrycks2021measuring}, TruthfulQA~\cite{lin2022truthfulqa}, GSM8K~\cite{cobbe2021training}, MATH~\cite{hendrycks2measuring}, or domain-specific retain tasks.
In image generation, it is commonly measured by IS~\cite{salimans2016improved}, FID~\cite{heusel2017gans}, KID~\cite{bińkowski2018demystifying}, LPIPS~\cite{zhang2018unreasonable}, SSIM~\cite{wang2004image}, CLIP Score~\cite{hessel2021clipscore}, CLIP similarity~\cite{radford2021learning}, aesthetic scores~\cite{schuhmann2022laion}, TIFA-style alignment scores~\cite{hu2023tifa}, VQA-based image-text consistency~\cite{lin2024evaluating}, and human studies on retain prompts.
In audio generation, it includes Naturalness MOS, Similarity MOS, WER, speaker verification on retained speakers, acoustic similarity, and downstream audio-task accuracy~\cite{kim2025do}.
In multimodal generation, it includes captioning metrics, VQA accuracy, image-text retrieval, cross-modal alignment, multimodal reasoning benchmarks such as MMMU~\cite{yue2023mmmu}, and text-only knowledge retention~\cite{huo2025mmunlearner,li2024single}.
These metrics support average distribution preservation, but may hide severe damage in small target-adjacent regions.

\paragraph{Boundary preservation}
Boundary evidence evaluates whether retainable content near the target remains intact.
For instance-level events, boundary probes include near-duplicate but non-target samples.
For identity-level events, they include visually, acoustically, or semantically similar non-target identities, aliases, roles, or attributes.
For concept/style-level events, they include neighboring concepts, related styles, benign uses of sensitive topics, and visually or semantically close categories~\cite{amara2025erasebench,moon2024holistic}.
For relation/association-level events, they include the endpoints of the removed relation and nearby relations that should remain available.
For behavior/policy-level events, they include benign requests close to prohibited behaviors, such as educational safety discussion, legitimate medical or legal information, or valid tool invocations near restricted ones~\cite{ji2024pku,chen2024wpn}.
Boundary preservation operationalizes over-unlearning.
A method that lowers target risk but damages this slice shifts the error from leakage to excessive deletion.

Furthermore, the distance on a preservation slice can be instantiated by a task-performance gap, output-distribution divergence, embedding distance, perceptual distance, identity-verification gap, reward-model difference, or human/LLM judgment~\cite{tang2024learn}.
Its role is to measure how far $Q(\cdot|x)$ deviates from the original behavior on a specified preservation slice.
Thus, $\widehat D_{\mathrm{ret}}$ and $\widehat B_T$ mainly differ in where the distance is evaluated.
Broad retain preservation estimates average non-target stability; boundary preservation estimates local specificity near the target event.

\subsection{Operator-Cost Evidence}
\label{sec:operator_cost_implementation}

Operator cost should be reported as a cost profile rather than a single runtime number.
In GenMU, the cost of an unlearning operator may appear at different stages, including training, data construction, memory usage, parameter storage, inference, auxiliary model calls, and model access.
This stage-wise view is necessary because different operators shift cost in different ways.
Objective-reshaping operators usually incur optimization, retain-data, and checkpointing costs.
Preference or contrastive steering operators further require preference-pair construction, annotation, reference models, or reward models.
Localization-editing operators may reduce the final update size, but they often require attribution, causal tracing, saliency estimation, or representation analysis before editing.
Model-arithmetic or modular operators may pay an upfront cost to construct directions or modules, while amortizing this cost across repeated or compositional requests~\cite{jia2024soul}.
Interface or decoding-control operators may avoid training, but they introduce prompt rewriting, retrieval filtering, detector calls, logit guidance, or inference-time latency.
Therefore, operator cost depends on the pair $(\mathcal{T},\Omega)$ and the deployment setting.
Comparing methods only by training time is misleading when a method shifts cost to inference, assumes stronger access, relies on unreported auxiliary models, or requires additional retain or preference data.
Existing efficiency studies mainly instantiate this evidence through tuning-based versus RLHF-style time comparisons~\cite{chen2023unlearn,yao2024large}, detailed wall-clock profiling~\cite{shi2024muse}, and amortized modular overhead~\cite{jia2024soul}.
A fair evaluation should therefore report the main cost stages of $\Omega$, rather than reducing cost to a single wall-clock number.

\subsection{Evidence Validity}
\label{sec:evidence_validity}

The previous subsections describe common evidence construction forms, but they do not by themselves guarantee a valid unlearning claim.
A finite evaluation supports the intended claim only when it satisfies three conditions: coverage, alignment, and resolution.
These conditions bridge evidence construction and failure diagnosis.
\begin{enumerate}[leftmargin=*]
    \item \textbf{Coverage.} Coverage asks whether $\widehat{\mathfrak{Q}}_T$ and $\widehat{\mathfrak{P}}_T$ cover the relevant regions of the target event.
    On the target side, weak coverage yields fragile suppression: the target disappears under evaluated queries but survives under untested transformations.
    On the preservation side, weak coverage yields hidden distortion: the model appears useful on broad benchmarks but fails on target-adjacent retainable content.
    Coverage is therefore not just sample size.
    It asks whether finite probes span the access paths and preservation slices.
    \item \textbf{Alignment.} Alignment asks whether $\widehat h_T$ and the preservation distance measure the intended semantic quantities.
    A target presence criterion should detect the target event rather than a loose correlate.
    A preservation distance should reflect behavior that should remain stable, not superficial output similarity alone.
    Misalignment is common because generative outputs are open-ended.
    Low lexical overlap may miss paraphrased leakage; a toxicity classifier may over-penalize benign discussion; a CLIP score may conflate target removal with prompt-image misalignment; a face verifier may overreact to style changes; and an LLM judge may introduce its own preference bias.
    Thus, estimator alignment must be justified by the target event and preservation slice, not inherited from modality convention.
    \item \textbf{Resolution.} Resolution asks whether the evaluation separates broad and boundary preservation.
    If only $\widehat D_{\mathrm{ret}}$ is reported, severe boundary damage can be averaged away.
    If only $\widehat B_T$ is reported, broad utility collapse may be missed.
    A credible evaluation therefore requires both slices.
    This distinction is especially important for identity-level, concept/style-level, relation/association-level, and behavior/policy-level target events, where the target boundary is dense and ambiguous.
\end{enumerate}
Together, coverage, alignment, and resolution determine whether the evidence vector is a reliable certificate or merely a collection of scores.
Insufficient coverage produces hidden target leakage.
Poor alignment produces false suppression or false damage estimates.
Low resolution hides the distinction between average preservation and target-adjacent preservation.
These defects lead directly to the diagnostic failures.

\begin{table*}[t]
\centering
\caption{Diagnostic map of GenMU failure modes. $\checkmark$ denotes the primary diagnostic anchor or evidence defect; $\star$ denotes a secondary one; -- denotes a not central one.}
\label{tab:diagnostic_map}
\scriptsize
\setlength{\tabcolsep}{2.8pt}
\renewcommand{\arraystretch}{1.12}
\resizebox{\linewidth}{!}{
\begin{tabular}{p{0.17\linewidth} ccccc cccc p{0.21\linewidth} p{0.22\linewidth}}
\toprule
\multirow{2}{*}{\textbf{Failure Mode}}
& \multicolumn{5}{c}{\textbf{Diagnostic Anchor}}
& \multicolumn{4}{c}{\textbf{Evidence Defect}}
& \multirow{2}{*}{\textbf{Signature}}
& \multirow{2}{*}{\textbf{Required Evidence}} \\
\cmidrule(lr){2-6}
\cmidrule(lr){7-10}
& $\widehat{R}_T$
& $\widehat{R}_T^{\mathrm{rob}}$
& $\widehat{D}_{\mathrm{ret}}$
& $\widehat{B}_T$
& $C_{\Omega}$
& Cov.
& Align.
& Res.
& Cost
& & \\
\midrule

Under-unlearning
& \cmark & \smark & \dash & \dash & \dash
& \dash & \smark & \dash & \dash
& Target remains under standard queries.
& Direct target probes with aligned $\widehat h_T$. \\

Fragile suppression
& \dash & \cmark & \dash & \dash & \dash
& \cmark & \dash & \dash & \dash
& Standard probes pass, transformed probes fail.
& Transformed, adversarial, multi-hop, and cross-modal queries. \\

Rerouting instead of erasing
& \dash & \cmark & \dash & \dash & \dash
& \cmark & \smark & \dash & \dash
& Direct outputs vanish, alternative paths recover target.
& Reverse queries, probability probes, multi-hop tests, and recovery prompts. \\

Relearning and recovery
& \dash & \cmark & \dash & \dash & \dash
& \cmark & \dash & \dash & \dash
& Suppression degrades after adaptation or retrieval changes.
& Prompt search, adapters, light updates, and retrieval-augmented recovery. \\

Over-unlearning
& \dash & \dash & \dash & \cmark & \dash
& \dash & \smark & \cmark & \dash
& Target is suppressed, but nearby retainable content is damaged.
& Boundary probes over target-adjacent retainable content. \\

Utility collapse
& \dash & \dash & \cmark & \dash & \dash
& \cmark & \dash & \smark & \dash
& Broad non-target behavior degrades.
& Retain benchmarks, retain-set performance, and generation-quality evidence. \\

Cross-modal leakage
& \dash & \cmark & \dash & \dash & \dash
& \cmark & \smark & \dash & \dash
& One modality passes, another modality leaks.
& Cross-modal transformations and modality-specific $\widehat h_T$. \\

Metric misalignment
& \smark & \smark & \smark & \smark & \dash
& \dash & \cmark & \dash & \dash
& Scores conflict with qualitative or adversarial evidence.
& Estimator validation, complementary metrics, and human/LLM audits. \\

Cost shifting
& \dash & \dash & \dash & \dash & \cmark
& \dash & \dash & \dash & \cmark
& Low reported cost hides auxiliary or inference overhead.
& Stage-wise $C_{\Omega}$ reporting. \\

Non-auditable unlearning
& \smark & \smark & \smark & \smark & \smark
& \smark & \smark & \smark & \smark
& Scores cannot be traced to $T$, $\mathcal{I}_T$, $\Omega$, or $C_{\Omega}$.
& Complete reporting of $T$, $\mathcal{I}_T$, $\Omega$, $\mathcal{Q}_{\Omega}$, and $C_{\Omega}$. \\

\bottomrule
\end{tabular}
}

\vspace{2pt}

\footnotesize
Cov.: coverage; Align.: alignment; Res.: resolution; Cost: cost reporting.
The Diagnostic Anchor columns specify which evidence-vector component reveals the failure.
The Evidence Defect columns specify which evaluation-design defect usually makes the failure hidden or non-auditable.

\vspace{-8pt}
\end{table*}

\subsection{Diagnostic Map of Failure Modes}
\label{sec:diagnostic_map}

The validity conditions above determine whether an evidence vector can support a GenMU claim.
When coverage, alignment, resolution, or cost reporting is insufficient, the reported evidence no longer certifies the claimed behavior.
The resulting mismatch appears as a diagnostic failure.
Insufficient coverage leaves target access paths or preservation slices untested, leading to fragile suppression, recovery, or cross-modal leakage.
Poor alignment makes the target presence criterion or preservation distance measure a proxy rather than the intended quantity, leading to metric misalignment or false conclusions about suppression and preservation.
Low resolution conflates broad retain behavior with target-adjacent behavior, making utility collapse or over-unlearning difficult to diagnose.
Incomplete cost reporting hides where the operator actually spends computation, memory, data, access, or inference resources.
Thus, diagnostic failures are not separate issues from evaluation; they are observable consequences of defective evidence design.

Tab.~\ref{tab:diagnostic_map} summarizes this connection.
The Diagnostic Anchor columns indicate which component of the evidence vector exposes the failure.
For example, high $\widehat{R}_T$ exposes under-unlearning, high $\widehat{R}_T^{\mathrm{rob}}$ exposes fragile suppression or recovery, high $\widehat{B}_T$ exposes over-unlearning, and incomplete $C_{\Omega}$ exposes cost shifting.
The evidence defect columns explain why the failure may be missed in the first place: the probes may not cover the relevant region, the estimator may not align with the target event, the evaluation may lack boundary resolution, or the cost profile may be incomplete.
The final two columns give the observable signature and the minimal additional evidence needed to verify or reject the corresponding claim.
Therefore, the table should be read as a diagnostic guide from evidence defect to failure mode, rather than as a list of independent evaluation metrics.

The diagnostic map leads to three practical rules.
First, target suppression cannot be claimed from $\widehat{R}_T$ alone.
A low $\widehat{R}_T$ only shows suppression under standard target queries, while fragile suppression, rerouting, relearning, and cross-modal leakage require transformed, adversarial, recovery-oriented, or cross-modal probes.
Second, distribution preservation cannot be claimed from $\widehat{D}_{\mathrm{ret}}$ alone.
A low $\widehat{D}_{\mathrm{ret}}$ supports average retain behavior, but over-unlearning remains possible unless $\widehat{B}_T$ is evaluated on target-adjacent retainable content.
Third, operator feasibility cannot be claimed from a single runtime number.
A low reported training time does not characterize $C_{\Omega}$ unless data construction, access assumptions, memory, auxiliary components, parameter storage, and inference overhead are also disclosed.
These rules turn the evidence vector into an audit tool: each GenMU claim should be checked against the evidence component that can falsify it and the evidence defect that may hide its failure.
\section{Applications Revisited through the Unified Unlearning Framework}
\label{sec:applications}

Sections~\ref{sec:target_taxonomy}--\ref{sec:evaluation} establish a unified view of GenMU: a deployment request first induces a target event $T$, then requires an unlearning operator $\Omega$ to transform the model distribution under preservation and cost constraints.
This section revisits major applications under this view.
The goal is not to introduce new application-specific taxonomies.
Instead, privacy protection, copyright protection, safety alignment, hallucination mitigation, and deployment defense can all be understood as different instantiations of target events and unlearning operators.
This perspective prevents application labels from obscuring the actual technical problem.
For example, privacy protection may involve memorized instances, identities, and private relations; copyright protection may involve protected samples, styles, personas, and author-work associations; safety alignment may involve harmful concepts and policy-violating behaviors.
Thus, an application is not itself a target-event category.
It is a composition of target structures, operator choices, and deployment constraints.


\begin{table*}[htbp]
\centering
\caption{Applications of GenMU as instantiations of target events and unlearning operators. $\checkmark$ denotes a dominant instantiation, $\star$ denotes a secondary or deployment-dependent one, and --: denotes a not central one.}
\label{tab:application_framework}
\scriptsize
\setlength{\tabcolsep}{2.6pt}
\renewcommand{\arraystretch}{1.12}
\resizebox{\linewidth}{!}{
\begin{tabular}{l ccccc ccccc p{0.26\linewidth}}
\toprule
\multirow{2}{*}{\textbf{Application}}
& \multicolumn{5}{c}{\textbf{Target-Event Instantiation}}
& \multicolumn{5}{c}{\textbf{Unlearning-Operator Instantiation}}
& \multirow{2}{*}{\textbf{Main Deployment Tension}} \\
\cmidrule(lr){2-6}
\cmidrule(lr){7-11}
& \textbf{Inst.} & \textbf{Id.} & \textbf{Con.} & \textbf{Rel.} & \textbf{Beh.}
& \textbf{Obj.} & \textbf{Pref.} & \textbf{Loc.} & \textbf{Mod.} & \textbf{Int./Dec.}
& \\
\midrule

\mbox{Privacy protection}
& \cmark & \cmark & \dash & \cmark & \dash
& \cmark & \dash & \cmark & \smark & \cmark
& Suppress private information while preserving benign public or domain knowledge \\

\mbox{Copyright and style protection}
& \cmark & \smark & \cmark & \cmark & \dash
& \cmark & \dash & \cmark & \cmark & \smark
& Suppress protected expression without collapsing neighboring creative ability \\

\mbox{Safety and preference alignment}
& \dash & \dash & \cmark & \dash & \cmark
& \cmark & \cmark & \smark & \dash & \cmark
& Suppress unsafe behavior while preserving helpful benign responses \\

\mbox{Hallucination mitigation}
& \dash & \dash & \dash & \cmark & \cmark
& \cmark & \dash & \smark & \dash & \cmark
& Remove false associations without damaging true neighboring knowledge \\

\mbox{Attack, defense, and audit}
& \smark & \dash & \smark & \smark & \cmark
& \cmark & \dash & \cmark & \cmark & \cmark
& Block exploitability without shifting risk to hidden access paths or deployment cost \\

\bottomrule
\end{tabular}
}

\vspace{2pt}

\footnotesize
Inst.: instance-level; Id.: identity-level; Con.: concept/style-level; Rel.: relation/association-level; Beh.: behavior/policy-level.
Obj.: objective reshaping; Pref.: preference or contrastive steering; Loc.: localization editing; Mod.: model-arithmetic or modular intervention; Int./Dec.: interface or decoding control.
\vspace{-8pt}
\end{table*}

\subsection{Privacy Protection}
\label{sec:privacy_protection}

Privacy protection is a central motivation for GenMU because GMs may reproduce private strings, personal documents, user profiles, speaker identities, faces, or sensitive associations observed during training.
In the unified framework, privacy is not a single target type.
It usually instantiates a composite target event:
\begin{equation}
    \mathcal{T}_{\mathrm{priv}}
    =
    \{
    \mathcal{T}_{\mathrm{inst}},
    \mathcal{T}_{\mathrm{id}},
    \mathcal{T}_{\mathrm{rel}}
    \},
\end{equation}
where instance-level events capture memorized private samples, identity-level events capture identifiable subjects, and relation-level events capture private associations such as person-phone-number or user-location relations.

This target structure determines the operator requirement.
If private content is available as explicit forget data, objective-reshaping operators can directly suppress its likelihood.
If private information is concentrated in identifiable representations, localization-editing operators can provide more scoped intervention.
If model access is restricted, interface or decoding-control operators can reduce leakage at deployment time, although such methods should be interpreted as access control unless recovery attempts fail.
Thus, the same privacy request can require different operators depending on whether the target is an instance, an identity, or a relation.

The main deployment tension is specificity.
The model should suppress private information without deleting public knowledge about the same domain or benign information about similar individuals.
For relation-level privacy, the endpoints of the private relation should remain usable while the relation itself is removed.
This is difficult because deleting one sensitive datum may expose another related datum, producing the onion effect~\cite{carlini2022privacy}.
Therefore, privacy-oriented GenMU should be framed as controlled suppression of a structured target event, not as indiscriminate removal of all information near a person, document, or domain.

\subsection{Copyright and Style Protection}
\label{sec:copyright_style_protection}

Copyright and style protection addresses unauthorized reproduction or imitation of protected content.
The protected target may be a memorized paragraph, an image, a character identity, an author-work association, or an artistic style.
Accordingly, this application instantiates instance-level, identity-level, relation-level, and concept/style-level target events.
A protected work is an instance-level event.
A recognizable character or persona is an identity-level event.
An artistic style is a concept/style-level event.
An author-work or image-caption link is a relation-level event.

This target structure makes boundary control central.
Copyright protection should suppress protected expression, but it should not remove generic visual composition, ordinary language patterns, public-domain concepts, or neighboring styles.
For this reason, localization-editing operators are attractive when protected content can be associated with cross-attention maps, latent directions, or concept-specific features.
Model-arithmetic or modular operators can support repeated style or concept removal by storing reusable directions or modules.
Objective-reshaping operators are useful when protected prompts or paired examples are available.
Interface or decoding-control operators can reduce immediate reproduction through prompt rewriting, negative prompts, retrieval filtering, or guidance, but they do not by themselves establish distributional removal.

This application also clarifies the relationship between GenMU and model editing.
Both may rely on locating and modifying target-related components, but their objectives differ.
Model editing usually corrects, inserts, or updates knowledge with minimal locality damage~\cite{yao2023editing,wang2024knowledge,huang2025diffusion}.
GenMU instead suppresses a target event while preserving surrounding distributional behavior.
The locate-then-modify paradigm is therefore shared~\cite{jung2025come}, but the desired direction of change is different.
For copyright and style protection, the core challenge is not only to remove protected outputs, but to avoid turning a legally scoped removal request into broad creative suppression.

\subsection{Safety and Preference Alignment}
\label{sec:safety_preference_alignment}

Safety and preference alignment targets harmful, biased, toxic, or policy-violating generations.
Large GMs trained on heterogeneous corpora may reproduce or amplify social and cultural biases, unsafe instructions, or harmful associations~\cite{chen2023ethics}.
Under the target-event taxonomy, these applications mainly instantiate concept/style-level and behavior/policy-level events.
A harmful concept, stereotype, or toxic attribute is a concept/style-level event.
Unsafe instruction following, jailbreak compliance, biased refusal, or tool misuse is a behavior/policy-level event.

This structure favors operators that act on relative behavior rather than isolated samples.
Preference or contrastive steering operators can suppress undesirable outputs by contrasting harmful and benign responses.
Objective-reshaping operators can directly penalize target behaviors when target prompts and retain data are available.
Interface or decoding-control operators can provide deployment-time safety through prompt guards, filters, reward-guided decoding, or expert/anti-expert steering.
These operators connect GenMU with RLHF and controllable generation.
RLHF optimizes reward signals to shape preferred outputs~\cite{kaufmann2023survey}, while GenMU removes or suppresses target events that should not be generated.
Controllable generation guides outputs through conditioning or decoding constraints~\cite{liang2024controllable,cao2024controllable}, while GenMU can be viewed as a negative control that excludes forbidden regions.

The main deployment tension is the boundary between harmful and benign behavior.
A safety operator may reduce harmful outputs by learning broad refusal, but this can damage helpfulness on benign adjacent requests.
For example, suppressing dangerous instructions should not prevent educational, medical, legal, or scientific discussion when it is safe and appropriate.
Thus, safety-oriented GenMU should not be judged only by a lower toxicity or harmfulness score.
It must preserve benign behavior near the policy boundary while remaining robust to transformed and adversarial access paths.

\subsection{Hallucination Mitigation}
\label{sec:hallucination_mitigation}

Hallucination mitigation targets outputs that are plausible but false, unsupported, or misleading.
In the unified framework, hallucination is not a single concept to erase.
It usually instantiates relation-level and behavior/policy-level target events.
A false entity-attribute association is a relation-level event.
A tendency to fabricate unsupported answers under uncertainty is a behavior/policy-level event.
This distinction matters because deleting a false relation and changing a hallucination behavior require different operators.

When hallucinations arise from memorized false data or spurious associations, GenMU can suppress the responsible target event.
Prior work shows that removing erroneous or fabricated knowledge can reduce hallucination rates~\cite{yao2024large}.
However, hallucinations also arise from reasoning limitations, retrieval failures, and uncertainty miscalibration.
In these cases, unlearning alone cannot solve the full problem.
Objective-reshaping operators may suppress specific false relations.
Localization-editing operators may modify representations associated with erroneous knowledge.
Retrieval control can prevent unreliable evidence from entering the generation context.
Model editing can complement GenMU when the goal is to replace a false relation with a correct one rather than simply suppress it~\cite{yao2023editing,wang2024knowledge,huang2025diffusion}.

The main deployment tension is correction versus deletion.
If the system only blocks a false answer, it may over-abstain on neighboring true questions.
If it inserts a correction without suppressing the false association, the old relation may reappear through multi-hop reasoning.
Therefore, hallucination mitigation should distinguish target relations from surrounding factual knowledge and should distinguish unsupported-answer behavior from legitimate uncertainty handling.
Under this view, GenMU is most useful when the hallucination can be localized as a target event; when hallucination reflects general reasoning weakness, it should be combined with retrieval, verification, or reasoning improvements.

\subsection{Attack, Defense, and Deployment Audit}
\label{sec:attack_defense_audit}

GenMU also supports defense against extraction, jailbreak, backdoor, and recovery attacks.
These attacks enlarge the effective query scope by searching for alternative access paths to private, unsafe, copyrighted, or backdoored content.
Under the unified framework, the corresponding targets may be memorized instances, private relations, or behavior/policy-level events.
The attacker is therefore not external to the target event.
It defines part of the transformation space through which the target may be recovered.

This setting favors operators designed for robustness and adaptation.
Adversarial unlearning combines targeted erasure with adversarial training, meta-learning, or bi-level optimization to reduce exploitation and relearning of removed knowledge~\cite{zhang2024defensive}.
Localization-editing operators can remove specific vulnerable components when the attack pathway is identifiable.
Model-arithmetic or modular operators can store reusable defense directions for repeated or compositional threats.
Interface or decoding-control operators are useful as deployment shields, but their protection is limited if prompt search or auxiliary modules can bypass the control layer.

The main deployment tension is auditability.
A defense claim is credible only when the target event, the unlearning operator, the reachable distribution family, and the cost profile are explicit.
Otherwise, a reported decrease in attack success rate may simply reflect a narrow attack set, a hidden filter, or unreported inference overhead.
This connects the application back to the diagnostic map in Section~\ref{sec:diagnostic_map}.
Rerouting, relearning, cross-modal leakage, and non-auditable unlearning are not separate deployment issues.
They are consequences of treating attack defense as a surface-level metric rather than as a target-event and operator instantiation problem.

\section{Open Problems and Future Directions}
\label{sec:future}

The preceding sections translate GenMU into a target-event and unlearning-operator problem.
This translation clarifies existing applications, but it also exposes what remains unresolved.
A mature GenMU system must specify the target event $T$, choose an unlearning operator $\Omega$, instantiate finite evidence $\mathcal{I}_T$, and justify the resulting claim under deployment constraints.
Current methods still fall short on each of these steps.
Some rely on retraining equivalence that is poorly suited to GMs.
Some define targets too narrowly or ignore target-adjacent preservation.
Some work only under white-box access.
Some degrade under repeated requests.
Some fail across modalities.
Some report metrics are difficult to compare or audit.
This section organizes these limitations into six open problems: formal guarantees beyond retraining equivalence, target boundary specification, black-box and API-only unlearning, continual and compositional unlearning, cross-modal certification and auditability, and benchmark standardization.

\begin{table*}[htbp]
\centering
\caption{Open problems under the unified GenMU framework. Each row marks the framework components most directly involved in the corresponding problem. The table identifies where each open problem enters the framework.}
\label{tab:future_directions}
\scriptsize
\setlength{\tabcolsep}{2.8pt}
\renewcommand{\arraystretch}{1.12}
\resizebox{\linewidth}{!}{
\begin{tabular}{p{0.23\linewidth} ccccc p{0.24\linewidth} p{0.30\linewidth}}
\toprule
\multirow{2}{*}{\textbf{Open Problem}}
& \multicolumn{5}{c}{\textbf{Framework Locus}}
& \multirow{2}{*}{\textbf{Core Risk}}
& \multirow{2}{*}{\textbf{Future Direction}} \\
\cmidrule(lr){2-6}
& $\mathcal{T}$
& $\Omega$
& $\mathcal{Q}_{\Omega}$
& $\mathcal{I}_T/\widehat{\mathbf e}$
& Audit
& & \\
\midrule

Formal guarantees beyond retraining equivalence
& \smark & \cmark & \cmark & \smark & \dash
& Parameter closeness fails to certify robust target suppression.
& Develop event-level distributional guarantees over $\mathcal{T}$, $\mathfrak{P}_T$, and $\mathcal{Q}_{\Omega}$. \\

Target boundary specification
& \cmark & \dash & \dash & \cmark & \smark
& Narrow targets leak, while broad targets over-delete.
& Build boundary-aware $h_T$, preservation scopes, and compositional target definitions. \\

Robust black-box and API-only unlearning
& \smark & \cmark & \cmark & \cmark & \smark
& Interface control may mask rather than remove targets.
& Design certified prompt, retrieval, filtering, and decoding operators with stated robustness scopes. \\

Continual and compositional unlearning
& \smark & \cmark & \cmark & \smark & \cmark
& Sequential requests accumulate distortion or interfere.
& Develop history-aware operators, module composition rules, and cumulative evidence tracking. \\

Cross-modal certification and auditability
& \cmark & \smark & \dash & \cmark & \cmark
& Targets persist through shared alignment pathways.
& Construct cross-modal probes, event-consistent criteria, and auditable evidence records. \\

Benchmark standardization
& \cmark & \smark & \smark & \cmark & \cmark
& Methods solve different target events under incomparable evidence.
& Standardize target events, evidence vectors, aggregation rules, and cost profiles. \\

\bottomrule
\end{tabular}
}

\vspace{2pt}

\footnotesize
$\mathcal{T}$: target event; $\Omega$: unlearning operator; $\mathcal{Q}_{\Omega}$: reachable distribution family;
$\mathcal{I}_T/\widehat{\mathbf e}$: finite instantiation and evidence vector; Audit: traceable reporting and benchmarking.
$\checkmark$: primary locus; $\star$: coupled or secondary locus; --: usually not central.

\vspace{-8pt}
\end{table*}

\subsection{Formal Guarantees Beyond Retraining Equivalence}
\label{sec:future_guarantees}

Classical machine unlearning often uses retraining equivalence as its reference.
For GenMU, this reference is no longer sufficient.
Retraining modern GMs is usually unavailable, and even when a retrained reference exists, closeness to that reference does not ensure that the target event is suppressed under transformed queries.
A model may differ substantially in parameters while behaving similarly on the target event, or it may appear close on standard tasks while still leaking the target through paraphrases or multimodal cues.
Thus, the central guarantee should move from parameter-level similarity to event-level distributional control.

Under the unified framework, a guarantee should specify which part of the projection problem is controlled.
It should relate the unlearning operator $\Omega$ and its reachable distribution family $\mathcal{Q}_{\Omega}$ to target suppression, distribution preservation, and operator cost.
A useful guarantee should therefore answer three questions.
First, does $Q\in\mathcal{Q}_{\Omega}$ suppress the target event over the declared query scope $\mathfrak{Q}_T$?
Second, does it preserve behavior over the preservation scope $\mathfrak{P}_T$?
Third, does it achieve this within an acceptable cost profile $C_{\Omega}$?
This reframes the formal objective from matching retraining to satisfying target-constrained distributional projection.

\subsection{Target Boundary Specification}
\label{sec:future_boundary}

Formal guarantees require a well-defined target event.
This immediately exposes a second open problem: how to specify the boundary between what should be suppressed and what should remain.
In the notation of this survey, the target event $T=(\mathfrak{Q}_T,\mathfrak{P}_T,h_T)$ is meaningful only when the target presence criterion $h_T$ and the preservation scope $\mathfrak{P}_T$ are aligned.
If $h_T$ is too narrow, transformed leakage is missed.
If $h_T$ is too broad, target-adjacent retainable content is treated as target content.
If $\mathfrak{P}_T$ lacks boundary slices, over-unlearning cannot be observed.

This problem is most severe when the target is not a single sample.
An identity may appear through names, faces, voices, aliases, and relations.
A style may share texture, color, or composition with neighboring styles.
A relation may need to be removed while its endpoints remain usable.
A safety policy may need to block harmful instructions while preserving benign educational or scientific discussion.
These cases show that target definition is not a preliminary detail, but a determinant of whether target suppression and distribution preservation are compatible.
Future work should therefore make boundary construction explicit.
One direction is to build preservation scopes from target-adjacent but retainable examples.
Another is to design graded target presence criteria that distinguish target content from neighboring content.
A further direction is to describe complex concepts compositionally.
GMs may encode high-level concepts through simpler components, so removing only the high-level label may leave subordinate elements that can be recombined~\cite{shumailov2024ununlearning}.
Boundary-aware and compositional target definitions are therefore necessary for precise GenMU.

\subsection{Robust Black-Box and API-Only Unlearning}
\label{sec:future_blackbox}

Even with a well-defined target event, the unlearning operator may be severely constrained.
Many deployed GMs are available only through APIs.
In such settings, users cannot inspect parameters, compute gradients, localize internal components, or verify how the base distribution changes.
The feasible operator family is therefore limited to prompts, retrieval control, filters, external memory changes, or decoding-time interventions.

This restriction changes the meaning of unlearning.
An interface filter may block direct target queries.
A retrieval filter may remove target passages from context.
A decoding rule may reduce target token probability.
These interventions can be useful in deployment, but they may leave the underlying target information intact.
The target may reappear under paraphrases, auxiliary contexts, prompt search, or cross-modal queries.
Thus, black-box unlearning should not claim model-level removal unless recovery-oriented evidence supports that claim.
Future work should develop black-box operators with explicit robustness scopes.
Certified prompt or input transformations could define which query transformations are covered.
Retrieval-control systems could tie memory deletion to a declared target event and preservation scope.
Adaptive probing could search for leakage under a query budget.
API-only GenMU should therefore report what kind of claim it supports: deployment-time suppression, robust access control, or evidence of distributional removal.
Without this distinction, black-box unlearning risks becoming non-auditable filtering.

\subsection{Continual, Streaming, and Compositional Unlearning}
\label{sec:future_continual}

Black-box constraints are not the only deployment challenge.
Real systems also receive unlearning requests over time.
Most current methods assume a fixed batch of targets, but legal, privacy, and platform requirements may require immediate or repeated removal.
Applying a batch method iteratively can accumulate retain distortion, expand boundary distortion, or weaken earlier removals.
Thus, continual GenMU is not merely faster unlearning.
It is a stability problem over a sequence of target-constrained transformations.

Let $\{\mathcal{T}_1,\ldots,\mathcal{T}_K\}$ be a sequence of target events.
A continual system produces
\begin{equation}
    P_{\boldsymbol{\theta}_0}
    \xrightarrow{\Omega_1}
    Q_1
    \xrightarrow{\Omega_2}
    Q_2
    \cdots
    \xrightarrow{\Omega_K}
    Q_K .
\end{equation}
Each step can affect previous target suppression, future preservation scopes, and the cost of later requests.
This creates two coupled risks.
The first is accumulation, where small distributional changes gradually damage utility.
The second is interference, where removing one target weakens or reverses another removal.
Future methods should make the history of unlearning explicit.
Objective-reshaping methods need regularizers that preserve previous suppressions and retained behavior.
Localization-editing methods need conflict detection across target-related subspaces.
Modular methods need algebraic rules for composing, merging, and removing modules.
Interface-control methods need auditable management of a growing set of filters and retrieval rules.
The evaluation should also become sequential, reporting cumulative target suppression, cumulative distribution preservation, and amortized operator cost.
This direction directly extends the scalability concerns observed in existing methods, where performance can degrade sharply as more targets are removed~\cite{jang2023knowledge}.

\subsection{Cross-Modal Certification and Auditability}
\label{sec:future_crossmodal_audit}

Continual unlearning becomes even harder in multimodal systems because target information can move across observation spaces.
A target suppressed in text may remain accessible through images.
A face blocked in image generation may be recovered by captioning or VQA.
A speaker identity weakened acoustically may persist through transcript-conditioned prompts.
These examples show that single-modality evidence cannot certify GenMU when the model contains shared alignment pathways.

Cross-modal certification should therefore treat modality changes as part of the target query scope.
The empirical instantiation $\mathcal{I}_T$ should include cross-modal probes, and $\widehat h_T$ should be modality-specific but event-consistent.
For identity-level targets, this means coordinating name matching, face recognition, speaker verification, and cross-modal identity matching.
For relation-level targets, this means testing both directions of an association, such as image-to-text and text-to-image recovery.
For behavior-level targets, this means testing whether a policy violation can be triggered through textual, visual, audio, or multimodal inputs.

Certification must also be auditable.
An unlearning claim should be traceable to $\mathcal{T}$, $\mathcal{I}_T$, $\Omega$, $\mathcal{Q}_{\Omega}$, and $C_{\Omega}$.
Without this trace, one cannot distinguish successful unlearning from narrow prompting, hidden filtering, or unreported auxiliary computation.
Future audit protocols should record target-event definitions, probe construction, operator assumptions, access level, auxiliary models, and stage-wise cost profiles.
This connects robustness, interpretability, and compliance: GenMU should be effective, but it should also be inspectable.

\subsection{Benchmark Standardization under the Proposed Protocol}
\label{sec:future_benchmark}

The previous problems all lead to the same practical obstacle: current benchmarks are difficult to compare.
Different studies instantiate different target events, use different query scopes, omit different preservation slices, and report different cost profiles.
Adding more metrics does not solve this problem.
A benchmark becomes useful only when it standardizes what the target is, how it is probed, what must be preserved, and what cost is reported.

A benchmark under this survey's protocol should specify four components.
First, it should define the target event $\mathcal{T}=(\mathfrak{Q}_T,\mathfrak{P}_T,h_T)$.
Second, it should provide a finite instantiation $\mathcal{I}_T$ with standard queries, transformed queries, broad retain slices, and boundary slices.
Third, it should describe the operator class $\Omega$ and the reachable distribution family $\mathcal{Q}_{\Omega}$.
Fourth, it should report the evidence vector $\widehat{\mathbf{e}}$ with explicit aggregation rules.
Future benchmarks should also include harder settings by design.
They should include composite target events, transformed queries, target-adjacent preservation slices, cross-modal probes, and stage-wise cost reporting.
They should distinguish deployment-time suppression from model-level removal.
They should also support continual requests.
Such benchmarks would turn GenMU evaluation from metric collection into auditable evidence construction.

\section{Conclusion}
\label{sec:cou}

This paper surveys recent advances in GenMU across language, image, audio, and multimodal generation.
Rather than organizing existing studies by modality or surface-level implementation, we present a unified analytical framework that views GenMU as a target-event-driven transformation of a generative distribution.
Under this framework, an unlearning request is specified by a target event, realized through an unlearning operator, and assessed by empirical evidence over target suppression, distribution preservation, and operator cost.

The main contribution of this survey is to turn a fragmented and rapidly expanding literature into a coherent analytical structure.
By separating target events from application scenarios, we clarify what is actually being suppressed and where its preservation boundary lies.
By organizing methods according to unlearning operators, we compare existing approaches through their distributional mechanisms rather than their modality-specific implementations.
By recasting evaluation as evidence design, we connect commonly used metrics to the claims they can support and expose why failures such as fragile suppression, over-unlearning, cross-modal leakage, and non-auditable unlearning arise.

This framework also provides a practical foundation for future GenMU research.
It shows that privacy protection, copyright and style protection, safety alignment, hallucination mitigation, and deployment defense are not isolated problem settings, but concrete instantiations of target events and unlearning operators.
It further identifies key directions for the field, including formal guarantees beyond retraining equivalence, precise target-boundary specification, robust black-box and continual unlearning, cross-modal certification, and benchmark standardization.
Overall, this survey aims to support the development of GenMU methods that are not only effective but also precise, scalable, robust, and auditable.







\bibliographystyle{ACM-Reference-Format}
\bibliography{acm/sample-base}

\appendix

\section{Detailed Catalog of GenMU Methods by Unlearning Operators}
\label{app:operator_catalog}

Section~\ref{sec:operator_taxonomy} organizes GenMU methods by the reachable distribution family induced by each unlearning operator.
This appendix provides a detailed method catalog under the same operator-level view.
Each table corresponds to one unlearning operator.
For each method, we report its target event, method name, year, venue, generative space, operator mechanism, short description, and main evaluation focus.
This preserves the coverage of existing literature while aligning the method catalog with the distributional taxonomy used in the main text.

\paragraph{Notation.}
We use \textsc{TS}, \textsc{DP}, and \textsc{OC} to denote target suppression, distribution preservation, and operator cost, respectively.
The target-event labels follow Section~\ref{sec:target_taxonomy}: instance-level, identity-level, concept/style-level, relation/association-level, and behavior/policy-level.
When a method contains multiple implementation details, we classify it by the mechanism that primarily determines its reachable distribution family $\mathcal{Q}_{\Omega}$.


\providecommand{\EvalTS}{\textsc{TS}}
\providecommand{\EvalDP}{\textsc{DP}}
\providecommand{\EvalOC}{\textsc{OC}}

\subsection{Objective-Reshaping Operators}
\label{app:objective_reshaping}

Objective-reshaping operators obtain the post-unlearning distribution by optimizing an erase-retain objective.
They include gradient-based forgetting, retain-regularized fine-tuning, distillation-based unlearning, second-order correction, diffusion-objective reshaping, and multimodal alignment-based unlearning.
Their main strength is flexibility, since the same objective template can be adapted to different target events.
Their main limitation is that target suppression and distribution preservation must be balanced within the same optimization process.

\begin{table*}[h]
\centering
\caption{Detailed catalog of objective-reshaping operators.}
\label{tab:app_objective_reshaping}
\scriptsize
\setlength{\tabcolsep}{3pt}
\renewcommand{\arraystretch}{1.12}
\resizebox{\linewidth}{!}{
\begin{tabular}{c l c c c l p{0.34\linewidth} c}
\toprule
\textbf{Target Event}
& \textbf{Method}
& \textbf{Year}
& \textbf{Venue}
& \textbf{Space}
& \textbf{Mechanism}
& \textbf{Short Description}
& \textbf{Eval.} \\
\midrule

\multirow{18}{*}{\makecell[c]{Instance-\\level}}
& GA~\cite{jang2023knowledge}
& 2023
& ACL
& Text
& Erase-loss optimization
& Maximizes target-sentence loss to reduce sample generation.
& \EvalTS;\EvalDP \\
& FPGA~\cite{feng2024fine}
& 2024
& EMNLP
& Text
& Selective erase loss
& Applies token-level selective forgetting for controlled removal.
& \EvalTS;\EvalDP \\
& SGA~\cite{buarbulescu2024each}
& 2024
& ICML
& Text
& Selective erase loss
& Performs sequence-level selective gradient ascent.
& \EvalTS;\EvalDP \\
& SEUL~\cite{wang2025selective}
& 2025
& AAAI
& Text
& Span-level erase loss
& Suppresses target spans while preserving non-target behavior.
& \EvalTS;\EvalDP \\
& POP~\cite{lee2024protecting}
& 2024
& ACL
& Text
& Gradient-difference objective
& Combines target ascent with first-order gradient difference.
& \EvalTS;\EvalDP \\
& Gu et al.~\cite{gu2024second}
& 2024
& arXiv
& Text
& Fisher-aware objective
& Uses Fisher information to stabilize erase-retain updates.
& \EvalTS;\EvalDP;\EvalOC \\
& Wei et al.~\cite{wei2025provable}
& 2025
& ICLR
& Text
& Hessian-aware objective
& Uses Hessian correction for controlled unlearning.
& \EvalTS;\EvalDP;\EvalOC \\
& KGA~\cite{wang2023kga}
& 2023
& ACL
& Text
& Distillation objective
& Preserves retained behavior through knowledge distillation.
& \EvalDP;\EvalTS \\
& E2URec~\cite{wang2025towards}
& 2025
& FOCS
& Text
& Distillation objective
& Uses efficient distillation for recommendation unlearning.
& \EvalDP;\EvalOC \\
& $O^3$~\cite{gao2025large}
& 2025
& ICLR
& Text
& OOD-guided objective
& Uses OOD detection to isolate removal targets.
& \EvalTS;\EvalOC \\
& Premptis et al.~\cite{premptis2025ails}
& 2025
& arXiv
& Text
& Partition-guided objective
& Uses data partitioning to localize target influence.
& \EvalTS;\EvalOC \\
& LINGTEA~\cite{choi2024cross}
& 2024
& EMNLP
& Text
& Erase-retain KL objective
& Combines target ascent with retain-side KL control.
& \EvalTS;\EvalDP \\
& UOE~\cite{zhuang2024uoe}
& 2024
& arXiv
& Text
& Attribution-guided objective
& Uses expert attribution to guide target suppression.
& \EvalTS;\EvalDP \\
& Hotfixing~\cite{yang2024hotfixing}
& 2024
& arXiv
& Text
& Erase-retain KL objective
& Performs targeted hotfixing with retain regularization.
& \EvalTS;\EvalDP;\EvalOC \\
& Kong et al.~\cite{kong2023data}
& 2023
& SaTML
& Image
& Adversarial label objective
& Treats target samples as fake for generator/discriminator updates.
& \EvalTS;\EvalDP \\
& Sun et al.~\cite{sun2025generative}
& 2025
& TDSC
& Image
& Substitute-label objective
& Uses substitute images and cascaded sample/class removal.
& \EvalTS;\EvalDP \\
& Feng et al.~\cite{feng2025controllable}
& 2025
& ICLR
& Image
& Bi-objective optimization
& Combines embedding comparison and noisy perturbation objectives.
& \EvalTS;\EvalDP \\
& Li et al.~\cite{li2024machine}
& 2024
& ICLR
& Image
& KL-based objective
& Uses dual KL terms for forget and retain distributions.
& \EvalTS;\EvalDP \\

\midrule
\multirow{2}{*}{\makecell[c]{Identity-\\level}}
& Mason-Williams et al.~\cite{mason-williams2025machine}
& 2025
& -
& Audio
& Speaker-identity objective
& Removes speaker-specific information from speech synthesis.
& \EvalTS;\EvalDP \\
& TGU~\cite{kim2025do}
& 2025
& ICML
& Audio
& Teacher-guided objective
& Uses randomized teacher speech as negative supervision.
& \EvalTS;\EvalDP \\

\midrule
\multirow{28}{*}{\makecell[c]{Concept/\\style-level}}
& Eldan et al.~\cite{eldan2023s}
& 2023
& arXiv
& Text
& Alternative-label objective
& Fine-tunes on surrogate labels to erase a concept.
& \EvalTS;\EvalDP \\
& Liu et al.~\cite{liu2024revisiting}
& 2024
& EMNLP
& Text
& Causal objective
& Treats the concept as a confounder to be suppressed.
& \EvalTS;\EvalDP \\
& MOLLM~\cite{pan2025multi}
& 2025
& ICASSP
& Text
& Multi-objective optimization
& Balances concept removal and capability preservation.
& \EvalTS;\EvalDP \\
& Yao et al.~\cite{yao2024large}
& 2024
& NeurIPS
& Text
& Mismatch-and-KL objective
& Combines mismatch and KL losses for concept suppression.
& \EvalTS;\EvalDP \\
& PCGU~\cite{yu2023unlearning}
& 2023
& ACL
& Text
& Attribution-guided objective
& Uses parameter attribution for concept-level forgetting.
& \EvalTS;\EvalDP \\
& Bae et al.~\cite{bae2023gradient}
& 2023
& ICML
& Image
& Hessian-aware objective
& Applies second-order correction for visual concept removal.
& \EvalTS;\EvalDP;\EvalOC \\
& Moon et al.~\cite{moon2024feature}
& 2024
& AAAI
& Image
& Latent-projection objective
& Projects away concept-related latent features.
& \EvalTS;\EvalDP \\
& Malnick et al.~\cite{malnick2024taming}
& 2024
& WACV
& Image
& Probability-measure update
& Adjusts probability measures to remove visual concepts.
& \EvalTS;\EvalDP \\
& ESD~\cite{gandikota2023erasing}
& 2023
& ICCV
& Image
& Diffusion score objective
& Erases concepts by altering unconditional diffusion behavior.
& \EvalTS;\EvalDP \\
& ACE~\cite{wang2025ace}
& 2025
& CVPR
& Image
& Diffusion score objective
& Uses concept-erasing objectives for stronger suppression.
& \EvalTS;\EvalDP \\
& Kumari et al.~\cite{kumari2023ablating}
& 2023
& ICCV
& Image
& Anchor-distribution objective
& Pulls generation toward anchor distributions.
& \EvalTS;\EvalDP \\
& DT~\cite{ni2023degeneration}
& 2023
& ACM MM
& Image
& Anchor-distribution objective
& Uses degeneration objectives for concept suppression.
& \EvalTS;\EvalDP \\
& SAFEGEN~\cite{li2024safegen}
& 2024
& ACM CCS
& Image
& Anchor-distribution objective
& Suppresses unsafe concepts while retaining safe generation.
& \EvalTS;\EvalDP \\
& SDD~\cite{kim2023towards}
& 2023
& ICML
& Image
& Self-distillation objective
& Distills target-free behavior into diffusion models.
& \EvalTS;\EvalDP \\
& SalUn~\cite{fan2023salun}
& 2024
& ICLR
& Image
& Saliency-aware objective
& Uses saliency-guided updates for less collateral damage.
& \EvalTS;\EvalDP \\
& HFI~\cite{kim2024safeguard}
& 2024
& ECCV
& Image
& Prompt-inversion objective
& Uses prompt inversion to suppress unsafe concepts.
& \EvalTS;\EvalDP \\
& Shirkavand et al.~\cite{shirkavand2025efficient}
& 2025
& CVPR
& Image
& Bi-level objective
& Mines neighbor concepts for efficient concept unlearning.
& \EvalTS;\EvalDP;\EvalOC \\
& Receler~\cite{huang2024receler}
& 2024
& ECCV
& Image
& Masked diffusion objective
& Uses mask-guided objectives for refined concept erasure.
& \EvalTS;\EvalDP \\
& RACE~\cite{kim2024race}
& 2024
& ECCV
& Image
& Adversarial objective
& Uses adversarial training for robust concept removal.
& \EvalTS;\EvalDP \\
& AdvUnlearn~\cite{zhang2024defensive}
& 2024
& NeurIPS
& Image
& Adversarial objective
& Defensively unlearns concepts under adversarial prompts.
& \EvalTS;\EvalDP \\
& Bui et al.~\cite{bui2024erasing2}
& 2024
& NeurIPS
& Image
& Adversarial objective
& Strengthens concept erasure with adversarial objectives.
& \EvalTS;\EvalDP \\
& AGE~\cite{bui2025fantastic}
& 2025
& ICLR
& Image
& Adversarial generation erasure
& Uses adversarial generation to suppress target concepts.
& \EvalTS;\EvalDP \\
& STEREO~\cite{srivatsan2025stereo}
& 2025
& CVPR
& Image
& Structured adversarial objective
& Uses structured adversarial losses for concept suppression.
& \EvalTS;\EvalDP \\
& SA~\cite{heng2023selective}
& 2023
& NeurIPS
& Image
& Continual erase-retain objective
& Uses selective amnesia for iterative concept removal.
& \EvalTS;\EvalDP \\
& MUNBa~\cite{wu2024munba}
& 2025
& ICCV
& Image
& Nash-bargaining objective
& Balances erasure and retention through bargaining.
& \EvalTS;\EvalDP \\
& DuMo~\cite{han2025dumo}
& 2025
& AAAI
& Image
& Skip-connection objective
& Uses skip-connection strategies for iterative concept removal.
& \EvalTS;\EvalDP \\
& SIU~\cite{li2024single}
& 2024
& NeurIPS
& Multimodal
& Masked-token objective
& Removes visual recognition skills with masked KL losses.
& \EvalTS;\EvalDP \\
& MMUnlearner~\cite{huo2025mmunlearner}
& 2025
& ACL
& Multimodal
& Fisher-aware objective
& Uses saliency masks and hierarchical visual losses.
& \EvalTS;\EvalDP;\EvalOC \\

\midrule
\makecell[c]{Relation/\\association-level}
& CLIPErase~\cite{yang2024cliperase}
& 2025
& ACL
& Multimodal
& Alignment objective
& Lowers forget-pair similarity while retaining alignments.
& \EvalTS;\EvalDP \\

\midrule
\makecell[c]{Behavior/\\policy-level}
& Chakraborty et al.~\cite{chakraborty2024crossmodal}
& 2024
& EMNLP
& Multimodal
& Cross-modal behavior objective
& Unlearns harmful behavior through language-component updates.
& \EvalTS;\EvalDP;\EvalOC \\
\bottomrule
\end{tabular}
}
\vspace{-8pt}
\end{table*}

\subsection{Preference or Contrastive Steering Operators}
\label{app:preference_contrastive}

Preference or contrastive steering operators reshape the relative likelihood of target-related and target-free outputs.
They are especially useful when the target event is easier to specify through comparisons than through an absolute target presence criterion.
Their effectiveness depends on whether the paired or contrastive data covers the target query scope and target boundary distribution.

\begin{table*}[h]
\centering
\caption{Detailed catalog of preference or contrastive steering operators.}
\label{tab:app_preference_contrastive}
\scriptsize
\setlength{\tabcolsep}{3pt}
\renewcommand{\arraystretch}{1.12}
\resizebox{\linewidth}{!}{
\begin{tabular}{c l c c c l p{0.34\linewidth} c}
\toprule
\textbf{Target Event}
& \textbf{Method}
& \textbf{Year}
& \textbf{Venue}
& \textbf{Space}
& \textbf{Mechanism}
& \textbf{Short Description}
& \textbf{Eval.} \\
\midrule
\multirow{3}{*}{\makecell[c]{Instance-\\level}}
& NPO~\cite{zhang2024negative}
& 2024
& COLM
& Text
& Negative preference steering
& Treats forget data as negative preference examples.
& \EvalTS;\EvalDP \\
& AltPO~\cite{mekala2025alternate}
& 2025
& COLING
& Text
& Positive-negative preference steering
& Combines negative target and positive retain feedback.
& \EvalTS;\EvalDP \\
& SimNPO~\cite{fan2024simplicity}
& 2024
& NeurIPS
& Text
& Negative preference steering
& Simplifies NPO and removes reference-model bias.
& \EvalTS;\EvalDP \\

\midrule
\multirow{5}{*}{\makecell[c]{Concept/\\style-level}}
& Quark~\cite{lu2022quark}
& 2022
& NeurIPS
& Text
& Quantized preference steering
& Uses quantized feedback for controllable suppression.
& \EvalTS;\EvalDP \\
& Opt-Out~\cite{choi2024opt}
& 2025
& ACL
& Text
& Regularized negative preference
& Adds Wasserstein regularization to NPO.
& \EvalTS;\EvalDP \\
& CRINGE~\cite{adolphs2023cringe}
& 2023
& ACL
& Text
& Token-level contrastive steering
& Uses contrastive signals to reduce undesired tokens.
& \EvalTS;\EvalDP \\
& DUO~\cite{park2024direct}
& 2024
& NeurIPS
& Image
& Paired-image contrastive steering
& Uses paired images with and without unsafe concepts.
& \EvalTS;\EvalDP \\
& Diffuse-DPO~\cite{wallace2024diffusion}
& 2024
& CVPR
& Image
& Diffusion preference steering
& Applies DPO-style supervision to diffusion outputs.
& \EvalTS;\EvalDP \\
\bottomrule
\end{tabular}
}
\vspace{-8pt}
\end{table*}

\subsection{Localization-Editing Operators}
\label{app:localization_editing}

Localization-editing operators first identify target-related internal components and then edit them.
The localized component may be a neuron set, parameter subset, representation direction, attention map, latent subspace, sparse feature, or cross-modal alignment pathway.
This family is suitable when the target event is concentrated in identifiable internal structures.
Its key requirement is that the localized component should cover the target event across the target query scope while avoiding target-adjacent retainable content.

\begin{table*}[h]
\centering
\caption{Detailed catalog of localization-editing operators.}
\label{tab:app_localization_editing}
\scriptsize
\setlength{\tabcolsep}{3pt}
\renewcommand{\arraystretch}{1.12}
\resizebox{\linewidth}{!}{
\begin{tabular}{c l c c c l p{0.34\linewidth} c}
\toprule
\textbf{Target Event}
& \textbf{Method}
& \textbf{Year}
& \textbf{Venue}
& \textbf{Space}
& \textbf{Mechanism}
& \textbf{Short Description}
& \textbf{Eval.} \\
\midrule
\multirow{4}{*}{\makecell[c]{Instance-\\level}}
& DEPN~\cite{wu2023depn}
& 2023
& EMNLP
& Text
& Attribution-based localization
& Locates target neurons by integrated-gradient attribution.
& \EvalTS;\EvalDP \\
& WAGLE~\cite{jia2025wagle}
& 2025
& NeurIPS
& Text
& Attribution-based localization
& Estimates parameter influence on forget and retain sets.
& \EvalTS;\EvalDP \\
& FALCON~\cite{hu2025falcon}
& 2025
& arXiv
& Text
& Information-theoretic localization
& Selects target components with information criteria.
& \EvalTS;\EvalDP \\
& SSU~\cite{dou2025avoiding}
& 2025
& NAACL
& Text
& Saliency-based localization
& Uses weight saliency for selective target suppression.
& \EvalTS;\EvalDP \\

\midrule
\multirow{13}{*}{\makecell[c]{Concept/\\style-level}}
& MemFlex~\cite{tian2024forget}
& 2024
& EMNLP
& Text
& Gradient-based localization
& Finds sensitive parameters for private/copyrighted content.
& \EvalTS;\EvalDP \\
& LAW~\cite{wang2025large}
& 2025
& ICLR
& Text
& Representation subspace editing
& Updates selected MLP layers for target redirection.
& \EvalTS;\EvalDP \\
& LUNAR~\cite{shen2025lunar}
& 2025
& arXiv
& Text
& Representation subspace editing
& Redirects forgotten data to non-answering regions.
& \EvalTS;\EvalDP \\
& Adaptive RMU~\cite{dang2025effects}
& 2025
& AAAI
& Text
& Layer-wise representation editing
& Adjusts intervention strength across layers.
& \EvalTS;\EvalDP \\
& UNLEARN~\cite{lizzo2024unlearn}
& 2025
& NAACL
& Text
& Subspace identification
& Identifies and edits target-related subspaces.
& \EvalTS;\EvalDP \\
& Farrell et al.~\cite{farrell2024applying}
& 2024
& NeurIPS
& Text
& Sparse-feature editing
& Intervenes on sparse target-related activations.
& \EvalTS;\EvalDP \\
& FMN~\cite{zhang2024forget}
& 2024
& CVPR
& Image
& Cross-attention localization
& Locates concept-specific attention regions.
& \EvalTS;\EvalDP \\
& MACE~\cite{lu2024mace}
& 2024
& CVPR
& Image
& Cross-attention editing
& Modifies attention modules for concept erasure.
& \EvalTS;\EvalDP \\
& UCE~\cite{gandikota2024unified}
& 2024
& WACV
& Image
& Closed-form attention editing
& Updates cross-attention projections in closed form.
& \EvalTS;\EvalDP \\
& RECE~\cite{gong2024reliable}
& 2024
& ECCV
& Image
& Robust attention editing
& Combines closed-form editing with adversarial training.
& \EvalTS;\EvalDP \\
& LocoGen~\cite{basu2024mechanistic}
& 2024
& ICML
& Image
& Mechanistic localization
& Uses causal localization for diffusion concept editing.
& \EvalTS;\EvalDP \\
& DiffQuickFix~\cite{basu2023localizing}
& 2024
& ICLR
& Image
& Causal mediation localization
& Uses mediation analysis to locate target pathways.
& \EvalTS;\EvalDP \\
& ConceptPrune~\cite{chavhan2025conceptprune}
& 2025
& ICLR
& Image
& Activation pruning
& Prunes concept-specific activations by importance.
& \EvalTS;\EvalDP \\

\midrule
\multirow{1}{*}{\makecell[c]{Identity-\\level}}
& MANU~\cite{liu2025modalityaware}
& 2025
& ACL
& Multimodal
& Modality-aware neuron localization
& Prunes target neurons across image-text pathways.
& \EvalTS;\EvalDP \\
\bottomrule
\end{tabular}
}
\vspace{-8pt}
\end{table*}

\subsection{Model-Arithmetic or Modular Operators}
\label{app:model_arithmetic_modular}

Model-arithmetic or modular operators represent unlearning through reusable directions, adapters, low-rank modules, safety vectors, expert/anti-expert directions, or external modules.
This family is attractive when unlearning requests are repeated, compositional, or deployment-oriented.
Its central assumption is that target-related behavior and retainable behavior are sufficiently separable in the chosen direction or module space.

\begin{table*}[h]
\centering
\caption{Detailed catalog of model-arithmetic or modular operators.}
\label{tab:app_model_arithmetic_modular}
\scriptsize
\setlength{\tabcolsep}{3pt}
\renewcommand{\arraystretch}{1.12}
\resizebox{\linewidth}{!}{
\begin{tabular}{c l c c c l p{0.34\linewidth} c}
\toprule
\textbf{Target Event}
& \textbf{Method}
& \textbf{Year}
& \textbf{Venue}
& \textbf{Space}
& \textbf{Mechanism}
& \textbf{Short Description}
& \textbf{Eval.} \\
\midrule
\makecell[c]{Instance-\\level}
& LoKU~\cite{cha2025towards}
& 2025
& ICLR
& Text
& Low-rank modular intervention
& Uses weighted low-rank modules for localized unlearning.
& \EvalTS;\EvalDP;\EvalOC \\

\midrule
\multirow{8}{*}{\makecell[c]{Concept/\\style-level}}
& Ilharco et al.~\cite{ilharco2023editing}
& 2023
& ICLR
& Text
& Task-vector arithmetic
& Uses weight-difference vectors to switch behaviors.
& \EvalTS;\EvalDP;\EvalOC \\
& Ethos~\cite{gao2024ethos}
& 2024
& NAACL
& Text
& Task-vector arithmetic
& Uses task-vector and SVD structure for debiasing.
& \EvalTS;\EvalDP;\EvalOC \\
& Zhou et al.~\cite{zhou2024making}
& 2024
& ACL
& Text
& Safety-vector intervention
& Applies reusable vectors to suppress unsafe behavior.
& \EvalTS;\EvalOC \\
& Ext-Sub~\cite{hu2024separate}
& 2024
& AAAI
& Text
& Expert-vector arithmetic
& Uses expert and anti-expert vector subtraction.
& \EvalTS;\EvalOC \\
& SKU~\cite{liu2024towards}
& 2024
& ACL
& Text
& Task-vector arithmetic
& Negates task vectors to remove target knowledge.
& \EvalTS;\EvalDP;\EvalOC \\
& SPM~\cite{lyu2024onedimensional}
& 2024
& CVPR
& Image
& Compact adapter intervention
& Uses one-dimensional adapters for concept suppression.
& \EvalTS;\EvalDP;\EvalOC \\
& Bui et al.~\cite{bui2024hiding}
& 2025
& ICLR
& Image
& Low-rank modular refinement
& Applies low-rank refinements to hide concepts.
& \EvalTS;\EvalDP;\EvalOC \\
& GLoCE~\cite{lee2025localized}
& 2025
& CVPR
& Image
& Localized low-rank module
& Uses localized low-rank modules for concept erasure.
& \EvalTS;\EvalDP;\EvalOC \\

\midrule
\makecell[c]{Behavior/\\policy-level}
& ToolDelete~\cite{cheng2025tool}
& 2025
& ICML
& Text
& Task-vector modular intervention
& Suppresses tool invocation with task-vector intervention.
& \EvalTS;\EvalDP;\EvalOC \\
\bottomrule
\end{tabular}
}
\vspace{-8pt}
\end{table*}

\subsection{Interface or Decoding-Control Operators}
\label{app:interface_decoding}

Interface or decoding-control operators suppress target events without necessarily modifying the base model.
They are especially useful when model parameters are unavailable or when unlearning must be deployed as a lightweight control layer.
Their limitation is that the underlying model distribution may remain unchanged, so target suppression should be tested over the full target query scope.

\begin{table*}[h]
\centering
\caption{Detailed catalog of interface or decoding-control operators.}
\label{tab:app_interface_decoding}
\scriptsize
\setlength{\tabcolsep}{3pt}
\renewcommand{\arraystretch}{1.12}
\resizebox{\linewidth}{!}{
\begin{tabular}{c l c c c l p{0.34\linewidth} c}
\toprule
\textbf{Target Event}
& \textbf{Method}
& \textbf{Year}
& \textbf{Venue}
& \textbf{Space}
& \textbf{Mechanism}
& \textbf{Short Description}
& \textbf{Eval.} \\
\midrule
\multirow{11}{*}{\makecell[c]{Instance-\\level}}
& ICUL~\cite{pawelczyk2024context}
& 2024
& ICML
& Text
& Prompt/context control
& Uses input context to simulate query-only unlearning.
& \EvalTS;\EvalOC \\
& ERASE~\cite{muresanu2024unlearnable}
& 2024
& arXiv
& Text
& Prompt/context control
& Selects prompts that steer away from forget content.
& \EvalTS;\EvalOC \\
& Thaker et al.~\cite{thaker2024guardrail}
& 2024
& arXiv
& Text
& Rule-based interface filtering
& Applies prompt or output filtering as guardrails.
& \EvalTS;\EvalOC \\
& SPUL~\cite{bhaila2025soft}
& 2025
& ACL
& Text
& Soft-prompt control
& Optimizes soft prompts for suppression and utility.
& \EvalTS;\EvalDP;\EvalOC \\
& ECO~\cite{liu2024large}
& 2024
& NeurIPS
& Text
& Prompt-classifier control
& Uses classifiers to control target-related prompts.
& \EvalTS;\EvalOC \\
& GRUN~\cite{ren2025general}
& 2025
& arXiv
& Text
& Gating-based interface control
& Uses gating for low-overhead API-side unlearning.
& \EvalTS;\EvalOC \\
& $\delta$-Unlearning~\cite{huang2024offset}
& 2025
& PMLR
& Text
& Logit-offset decoding control
& Applies estimated logit shifts for black-box suppression.
& \EvalTS;\EvalOC \\
& ULD~\cite{ji2024reversing}
& 2024
& NeurIPS
& Text
& Logit-offset decoding control
& Uses auxiliary LLMs to estimate output shifts.
& \EvalTS;\EvalOC \\
& ALU~\cite{sanyal2025alu}
& 2025
& COLM
& Text
& Agent-based decoding control
& Uses multiple agents to adjust target outputs.
& \EvalTS;\EvalOC \\
& MUNCH~\cite{choi2024breaking}
& 2024
& arXiv
& Text
& Uncertainty-aware decoding
& Uses uncertainty for controllable decoding-time unlearning.
& \EvalTS;\EvalOC \\
& CMC~\cite{wu2024cross}
& 2024
& NeurIPS
& Text
& Token-mapping decoding
& Applies token mapping for output-side control.
& \EvalTS;\EvalOC \\

\midrule
\multirow{14}{*}{\makecell[c]{Concept/\\style-level}}
& Wang et al.~\cite{wang2024machine}
& 2024
& arXiv
& Text
& Retrieval-context control
& Modifies retrieved context to simulate unlearning.
& \EvalTS;\EvalDP;\EvalOC \\
& DExperts~\cite{liu2021dexperts}
& 2021
& ACL
& Text
& Expert/anti-expert decoding
& Reweights tokens using expert and anti-expert LMs.
& \EvalTS;\EvalOC \\
& Li et al.~\cite{li2024get}
& 2024
& ICLR
& Image
& Embedding-space control
& Suppresses concept-bearing singular directions.
& \EvalTS;\EvalDP;\EvalOC \\
& POSI~\cite{wu2024universal}
& 2024
& NAACL
& Image
& Prompt-transformation control
& Uses reward guidance to transform unsafe prompts.
& \EvalTS;\EvalOC \\
& Ban et al.~\cite{ban2024understanding}
& 2024
& ECCV
& Image
& Negative-prompt control
& Uses negative prompts to suppress target concepts.
& \EvalTS;\EvalOC \\
& SLD~\cite{schramowski2023safe}
& 2023
& CVPR
& Image
& Diffusion-time safety guidance
& Applies safety guidance during denoising.
& \EvalTS;\EvalOC \\
& SEGA~\cite{brack2023sega}
& 2023
& NeurIPS
& Image
& Diffusion-time semantic guidance
& Uses semantic guidance to steer diffusion outputs.
& \EvalTS;\EvalDP;\EvalOC \\
& Prompt Slider~\cite{sridhar2025prompt}
& 2024
& ECCV
& Image
& Prompt-direction guidance
& Steers generation through paired prompt directions.
& \EvalTS;\EvalOC \\
& DAG~\cite{li2025detect}
& 2025
& CVPR
& Image
& Region-selective guidance
& Detects and guides target regions during generation.
& \EvalTS;\EvalDP;\EvalOC \\
& SAeUron~\cite{cywinski2025saeuron}
& 2025
& ICLR
& Image
& Sparse-feature decoding
& Uses sparse features for inference-time suppression.
& \EvalTS;\EvalDP;\EvalOC \\
& AdaVD~\cite{wang2025precise}
& 2025
& CVPR
& Image
& Value-space decoding
& Uses value-space decomposition for precise editing.
& \EvalTS;\EvalDP;\EvalOC \\
& Concept Algebra~\cite{wang2023concept}
& 2023
& NeurIPS
& Image
& Algebraic embedding control
& Edits concept directions algebraically in embedding space.
& \EvalTS;\EvalDP;\EvalOC \\
& FAST~\cite{panda2024fast}
& 2025
& TAI
& Image
& Prompt-filtering control
& Filters prompts by similarity to target concepts.
& \EvalTS;\EvalOC \\
& Li et al.~\cite{li2024get}
& 2024
& ICLR
& Image
& Embedding guidance
& Also serves as inference-time embedding control.
& \EvalTS;\EvalDP;\EvalOC \\
\bottomrule
\end{tabular}
}
\vspace{-8pt}
\end{table*}

\subsection{Notes on Classification}
\label{app:operator_classification_notes}

Some methods contain multiple mechanisms and may appear to belong to more than one operator family.
For consistency with Section~\ref{sec:operator_taxonomy}, we assign each method according to the mechanism that primarily determines its reachable distribution family $\mathcal{Q}_{\Omega}$.
For example, a method that localizes a subspace and then applies a low-rank intervention is classified as localization-editing if the reachable family is primarily constrained by the localized subspace, but as model-arithmetic or modular if the main intervention is a reusable low-rank module.
Similarly, a method that trains a soft prompt with an erase-retain objective is classified as interface or decoding-control when the base model is unchanged and the reachable family is induced by the prompt interface.
A method that uses preference data only to construct a decoding-time controller is classified as interface or decoding-control rather than preference steering, because the reachable family is induced by the decoding rule.
This rule preserves the distinction between implementation details and distributional mechanisms.

The catalog is not intended to be exhaustive.
Instead, it provides a structured map of representative GenMU methods under the operator taxonomy used in the main text.
Additional methods can be added by identifying the target event they address, the unlearning operator they instantiate, the reachable distribution family they induce, and the evaluation group most relevant to their claims.

\end{document}